%% file: main.tex
\documentclass{article}

\input{preamble}

\input{notation}

\input{macros}

\usepackage{xr}
\title{Multi-task Causal Learning with Gaussian Processes}

\author{%
	Virginia Aglietti \\
	University of Warwick\\
	The Alan Turing Institute\\
	\texttt{V.Aglietti@warwick.ac.uk} \\
	\And
	Theodoros Damoulas \\
	University of Warwick\\
	The Alan Turing Institute\\
	\texttt{T.Damoulas@warwick.ac.uk} \\		
	\And 
	Mauricio \'Alvarez\\
	University of Sheffield\\
	\texttt{Mauricio.Alvarez@sheffield.ac.uk} \\		
	\And 
	Javier Gonz\'alez\\
	Microsoft Research Cambridge\\
	\texttt{Gonzalez.Javier@microsoft.com} \\		
}

\begin{document}

\maketitle

\input{abstract}
\input{intro}
\input{motivation}

\input{related}
\input{background_problem}
\input{methodology}

\input{helicopter}
\input{experiments}

\input{conclusions}

\input{impact}

\bibliographystyle{apalike}
\bibliography{references}

\end{document}


\maketitle

\section{Proofs of theorems and additional theoretical results}
In this section we give the proofs for the theorems in the main text and an additional theoretical result regarding the minimality of the set $\C$.
  
\subsection{Proof of Theorem 3.1}
\begin{proof}
Consider a generic $\X_s \in \partsX$. $\ivalue^I_s$ and $\ivalue^N_s$ denote the values for the sets $\Xsbase^I_s$ and $\Xsbase^{N}_s$ respectively. $\cvalue = (\cvalue_s^I \cup \cvalue_s^N)$ represents the values for the set $\C^{N}$, $\cvalue_s^N$ is the value of $\C^{N}_s$ and  $\cvalue_s^I$ gives the value for $\C^{I}_s$. Notice that we can write the intervention on $\X_s$, that is $\DO{\X_s}{\x}$, as $\DO{\I^{I}_s}{\ivalue^I_s} \cup \DO{\X_s\backslash \I^{I}_s}{\x \backslash \ivalue^I_s}$. Any function $t_s(\x) \in \T$ can be written as:
\begin{align}
		t_s(\x) &= \expectation{}{Y|\DO{\X_s}{\x}} \nonumber \\
		&= \idotsint \expectation{}{Y|\DO{\I^I_s}{\ivalue^I_s},\DO{\X_s \backslash \I^I_s}{\x \backslash \ivalue^I_s},  \Xsbase^{N}_s = \ivalue^{N}_s, \C^{N}_s=\cvalue^{N}_s} \times \nonumber \\
		& \;\;\;\;\;\; \quad p(\ivalue^{N}_s, \cvalue^{N}_s|\DO{\X_s}{\x}) \text{d}\ivalue^{N}_s \text{d} \cvalue^{N}_s \nonumber \\	
		\nonumber \\
		& = \idotsint \expectation{}{Y|\DO{\I^I_s}{\ivalue^I_s},\DO{\X_s \backslash \I^I_s}{\x \backslash \ivalue^I_s},  \DO{\Xsbase^{N}_s}{\ivalue^{N}_s}, \C^{N}_s=\cvalue^{N}_s} \times \nonumber \\
		& \;\;\;\;\;\; \quad p(\ivalue^{N}_s, \cvalue^{N}_s|\DO{\X_s}{\x}) \text{d}\ivalue^{N}_s \text{d} \cvalue^{N}_s \;\;\;\;\;\;\;\;\;\;\;\;  \text{by} \;\;\; Y \indep \Xsbase^{N}_s|\X_s, \C^{N}_s \;\; \text{in} \;\; \graph_{\overline{\X_s} \underline{\Xsbase^{N}_s}}  \label{eq:first_step} \\
		\nonumber\\
		& = \idotsint \expectation{}{Y|\DO{\I^I_s}{\ivalue^I_s}, \DO{\Xsbase^{N}_s}{\ivalue^{N}_s}, \C^{N}=\cvalue^{N}_s} \times \nonumber \\
		& \;\;\;\;\;\; \quad p(\ivalue^{N}_s, \cvalue^{N}_s|\DO{\X_s}{\x}) \text{d}\ivalue^{N}_s \text{d} \cvalue^{N}_s \;\;\;\;\;\;\;\;\;\;\;\;  \text{by} \;\;\; Y \indep (\X_s \backslash \I^I_s)|\I, \C^{N}_s  \;\; \text{in}  \;\; \graph_{\overline{\I (\X_s \backslash \I^I_s)(\C^{N}_s)}} \label{eq:second_step} \\
		\nonumber 
\end{align}
		
\begin{align}
		& = \idotsint \expectation{}{Y|\DO{\I}{\ivalue},  \C^{N}_s=\cvalue^{N}_s} p(\ivalue^{N}_s, \cvalue^{N}_s|\DO{\X_s}{\x}) \text{d}\ivalue^{N}_s \text{d} \cvalue^{N}_s \nonumber \\
		\nonumber \\
		& = \idotsint \expectation{}{Y|\DO{\Xsbase}{\ivalue},  \C^{N}_s=\cvalue^{N}_s, \C^{I}_s=\cvalue^{I}_s} \times \nonumber\\
		& \quad  \;\;\;\;\; p(\cvalue^{I}_s|\DO{\Xsbase}{\ivalue},  \C^{N}_s=\cvalue^{N}_s)p(\ivalue^{N}_s, \cvalue^{N}_s|\DO{\X_s}{\x}) \text{d}\ivalue^{N}_s \text{d} \cvalue^{N}_s \text{d} \cvalue^{I}_s  \nonumber \\
		\nonumber\\
		& = \idotsint \expectation{}{Y|\DO{\Xsbase}{\ivalue},  \C^{N}=\cvalue} p(\cvalue^{I}_s| \C^{N}_s=\cvalue^{N}_s)p(\ivalue^{N}_s, \cvalue^{N}_s|\DO{\X_s}{\x})  \text{d}\ivalue^{N}_s \text{d} \cvalue^{N}_s \text{d} \cvalue^{I}_s \label{eq:inter_result} \\
		& \;\;\;\;\;\;\;\;\;\;\;\;  \text{by} \;\;\; \C^I_s \indep \I | \C^{N}_s  \;\; \text{in}  \;\; \graph_{\overline{\I}} \nonumber\\
		& = \idotsint \expectation{}{Y|\DO{\I}{\ivalue},  \C^N=\cvalue}  p(\cvalue^I_s|\cvalue^{N}_s)p(\ivalue^{N}_s, \cvalue^{N}_s|\DO{\X_s}{\x}) \text{d}\ivalue^{N}_s \text{d} \cvalue \nonumber \\		
		& = \idotsint \basefunction(\ivalue, \cvalue) p(\cvalue^I_s|\cvalue^{N}_s)p(\ivalue^{N}_s, \cvalue^{N}_s|\DO{\X_s}{\x}) \text{d}\ivalue^{N}_s \text{d} \cvalue  \label{eq:final_measure} 
\end{align}
where \eq \eqref{eq:first_step} follows from Rule 2 of \textit{do}-calculus while \eq \eqref{eq:second_step} and
\eq \eqref{eq:inter_result} follow from Rule 3 of \textit{do}-calculus \cite{pearl2000causality}. \eq \eqref{eq:final_measure} gives the causal operator. 
\end{proof}

\subsection{Proof of Corollary 3.1}
\begin{proof} 
Suppose there exists another set $\mat{A}$, different from $\parents(Y)$ and defined as $\mat{A} = \parents(Y) \backslash \parents(Y)_i$, where $\parents(Y)_i$ represents a single variable in $\parents(Y)$, such that \eq (2) holds for every set $\X_s$. This means that $\mat{A}$ blocks the front-door paths from all $\X_s \in \partsX$ to $Y$. That is, $\mat{A}$ also blocks the directed path from $\parents(Y) \in \partsX$ to $Y$ thus including descendants of $\parents(Y)$ which are ancestors of $Y$. This contradicts the definition of a parent as a variable connected to $Y$ through a direct arrow. The same reasoning hold for every set non containing all parents of $Y$ thus $\parents(Y)$ is the smallest set such that \eq (2) holds.
\end{proof}

\subsection{Proof of  Theorem 3.2}
\begin{proof} 
Suppose that $\C$ includes a node, say $C_i$, that has both an incoming and an outcoming unconfounded edge. The unconfounded incoming edge implies the existence of a set $\X_s$ for which $C_i$ is a collider on the confounded path from $\X_s$ to $Y$. At the same time, the unconfounded outcoming edge implies the existence of a set $\X_{s'}$ such that $C_i$ is an ancestor that we need to condition on in order to clock the back-door paths from $\X_{s'}$ to $Y$.  
Consequently, the conditions $Y \indep \Xsbase^{N}_s|\X_s, \C^{N}_s \;\; \text{in} \;\; \graph_{\overline{\X_s} \underline{\Xsbase^{N}_s}}$ and $Y \indep (\X_s \backslash \I^I_s)|\I, \C^{N}_s  \;\; \text{in}  \;\; \graph_{\overline{\I (\X_s \backslash \I^I_s)(\C^{N}_s)}}$ in Theorem 3.1 cannot hold, at the same time, for both $\X_s$ and $\X_{s'}$. Indeed, these independence conditions would be verified for $X_s$ when excluding $C_i$  from $\C^N$ while they would be verified for $X_{s'}$ when $C_i$ is included in $\C^N$. 
The same reasoning hold for every node in $\C$ having both incoming and outcoming unconfounded edges. Therefore, if $\graph$ has one of such node, it is not possible to find a set $\C$ such that \eq (2) holds from all $\X_s \in \partsX$. 
\end{proof}

\subsection{Additional corollary}\label{sec:additional_theorem}
\begin{corollary}
The set $\C$ represents the smallest set for which \eq (2) holds.
\end{corollary}

\begin{proof}
Suppose there exists another set $\mat{A}$, different from $\C$ and defined as $\mat{A} = \C \backslash C_i$ where $C_i \in \partsX$ denotes a single variable in $\C$ that is not a collider. The set $\mat{A}$ need to be such that $Y \indep (\X_s \backslash \I^I_s)|\I, \mat{A}^{N}_s  \;\; \text{in}  \;\; \graph_{\overline{\I (\X_s \backslash \I^I_s)(\mat{A}^{N}_s)}}$ 
\quad $\forall \X_s \hspace{0.2cm} \text{in} \hspace{0.2cm} \partsX$. Consider $\X_s = C_i$ and notice that the back door path from $C_i$ to $Y$ is \emph{not} blocked by conditioning on $\Xsbase$ or $\mat{A}^N_s$. Therefore $Y \not\!\perp\!\!\!\perp (\X_s \backslash \I^I_s)|\I, \mat{A}^{N}_s  \;\; \text{in}  \;\; \graph_{\overline{\I (\X_s \backslash \I^I_s)(\mat{A}^{N}_s)}}$ and $\mat{A}$ is not a valid set. The same reasoning holds for every set not containing all confounders of $Y$ thus $\C$ is the minimal set for $\C$. 
\end{proof}		

\section{Partial transfer}\label{sec:partial_transfer}
The conditions in Theorem 3.1 allow for full transfer across \emph{all} intervention functions in $\T$. 
As shown (see Theorem 3.2), this might not be possible when a subset $\C' \subset \C$ includes nodes directly confounded with $Y$ and with both unconfounded incoming and outcoming edges. However, we might still be interested in transferring information across a subset $\T^\prime \subset \T$ which includes functions defined on $\partsX^\prime \subset \partsX$. $\partsX^\prime$ is defined by excluding from $\partsX$ those intervention sets including variables that have outcoming edges pointing into $\C'$ making the conditions in Theorem 3.1 satisfied for all sets in $\partsX^\prime$. For instance, consider \fig 1 (b) with the red edge where $A$ is a confounded node that has both unconfounded incoming and outcoming edges. To block the path $E \leftarrow A \dashleftarrow \dashrightarrow Y$ we need to condition on $A$. However, conditioning on $A$ opens the path $F \rightarrow A \dashleftarrow \dashrightarrow Y $ making it impossible to define a base function. We can thus focus on a subset $\T' $ in which all functions including $\C' = \{A\}$ as an intervention variable have been excluded. This is equivalent to doing full transfer in \fig 1 (b) with no incoming red edge in $A$. 
	
\section{Advantages of using the Causal operator} 
\label{sec:operator}
The causal operator allows us to write any $t_s(\x)$ as an integral transformation of $\basefunction$. The integrating measure, which differ across $\X_s$, captures the dependency structure between the base set and the intervention set and can be reduced to \textit{do}-free operations via \textit{do}-calculus.
Notice how, given our identifiability assumptions, all functions in $\T$ can also be computed by simply applying the rules of \textit{do}-calculus when observational data are available. However, writing the functions via $L_s(\basefunction)(\x)$ has several advantages:
\begin{itemize}
	\item it allows to identify the correlation structure across functions and thus to specify a multi-task probabilistic model and share experimental information;
	\item it allows to learn those intervention functions for which we cannot run experiments via transfer;
	\item it allows to efficiently learn the set $\T$ when $\partsX$ is large.
\end{itemize}

This is crucial when have limited observational data or we cannot run experiments on some intervention sets or the cardinality of $\partsX$ is large.
In the last case, specifying a model for each individual intervention function would not only be computationally expensive but might also lead to inconsistent prior specification across functions. Through the causal operator we can model a system by only making one single assumption on $\basefunction$ which is then propagated in the causal graph. 
When an intervention is performed, the information is propagated in the graph through the base function which links the different interventional functions.  Using $\basefunction$ we avoid the specification of the correlation structure across every pair of intervention functions which would result in a combinatorial problem.


\section{Active learning algorithm} \label{sec:al_details}	
Denote by $D$  a set of inputs for the functions in $\T$, that is $D = \bigcup_s D_s$ with $D_s \subset D(\X_s)$ and consider a subset a set $A \subset D$ of size $k$. We would like to select $A$, that is select the both the functions to be observed and the locations, such that we maximize the reduction of entropy in the remaining unobserved locations:
\begin{align*}
A^\star =\argmax_{A:|A| = k} H(\T (D\backslash A) ) - H(\T (D\backslash A)|\T(A)) \text{.}
\end{align*}

where $\T (D\backslash A)$ denotes the set of functions $\T$ evaluated in $D\backslash A$, $\T (D\backslash A)|\T(A)$ gives the distribution for $\T$  at $(D\backslash A)$ given that we have observed $\T(A)$  while 
$H(\cdot)$ represents the entropy.
This problem is \acro{np}-complete, \citet{krause2008} proposed an efficient greedy algorithm providing an approximation for $A$. This algorithm starts with an empty set $A = \emptyset$ and solves the problem sequentially by selecting, at every step $j$, a point $\x_{sj} = \argmax_{ \x_{sj} \in D\backslash A} H(t_s(\x)|A) - H(t_s(\x)|D \backslash (A \cup  \x_{sj}))$. 
Both $H(t_s(\x)|A) = \frac{1}{2}\log(2\pi \sigma^2_{\x_{sj}|A})$ and $H(t_s(\x)|D \backslash (A \cup  \x_{sj})= \frac{1}{2}\log(2\pi \sigma^2_{\x_{sj}|D \backslash (A \cup  \x_{sj})})$ do not depend on the observed $\T$ values thus the set $A$ can be selected before any function evaluation is collected. For every $\X_s$,  $\sigma^2_{\x_{sj}|A}$ and $\sigma^2_{\x_{sj}|D \backslash (A \cup  \x_{sj})}$ correspond to the variance terms of the kernel on $\T$ and are thus determined by both the observational and the interventional data across all experiments. 

\begin{figure} 
	\centering
	\includegraphics[width=0.8\linewidth]{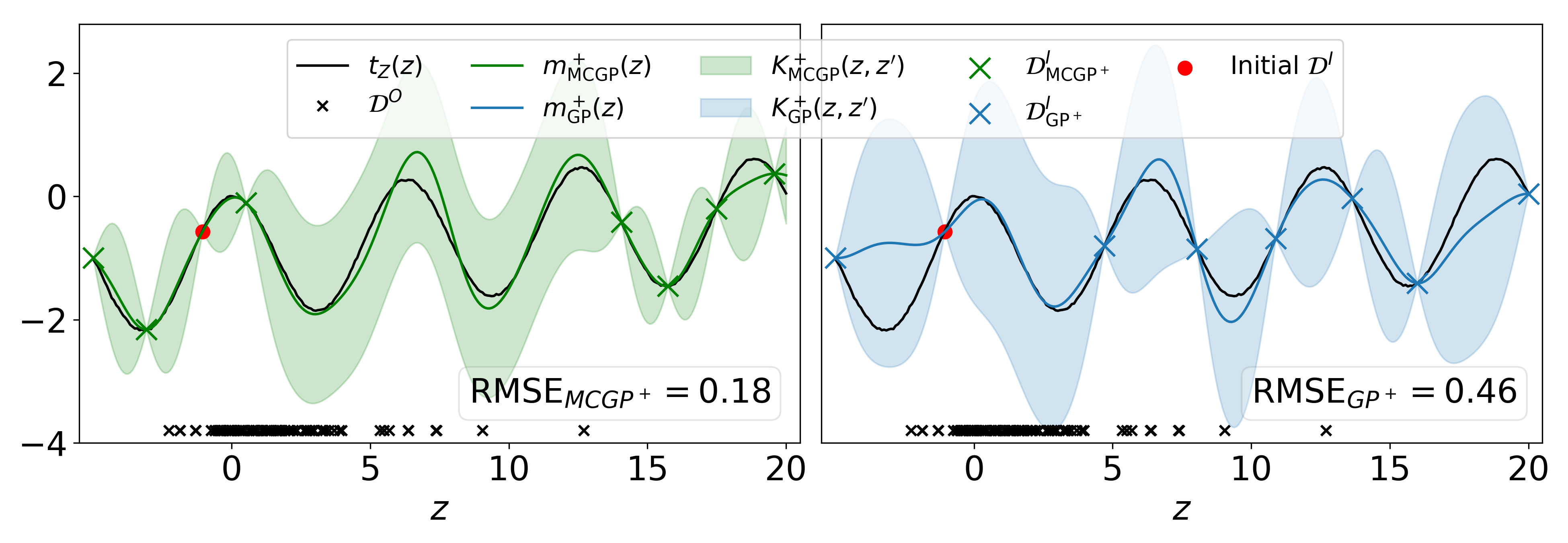} 
	\caption{Snapshot of the \al algorithm for $t_Z(z)$ of the \DAG in \fig 1 (a) when $N^I=8$ and $N=100$. $\our^+$ is our algorithm with the causal \gptext prior. $\gptext^+$ is a single-task model with the same prior (see \fig 3 for details on the compared models). Coloured crosses denote collected interventions while the red dot gives the common initial design.}
	\label{fig:al_toy_z}
\end{figure}

\fig \ref{fig:al_toy_z} shows a snapshot of the state of the \al algorithm for the toy example of \fig 1 (a) when 8 interventional data points have been collected for $t_Z(z)$. Both $\gptext^+$ and $\our^+$ avoid collecting data points in areas where the causal \gptext prior is already providing information thus making the model posterior mean equal to the true function (see region between $[0,5]$). $\gptext^+$ is spreading the function evaluations on the remaining part of the input space collecting data points in the region $[5,14]$. On the contrary, $\our^+$ drives the data points to be collected where neither observational nor interventional information can be transferred for the remaining tasks thus focusing on the border of the input space (see region $[14,20]$). Indeed, the variance structure for $t_Z(z)$ is reduced in $[5,14]$ (\fig \ref{fig:toy_1} (central panel)) by the interventional data collected for $t_X(x)$ when using $\our^+$ compared to $\gptext^+$.  Combining an \al framework with $\our^+$ is thus crucial when designing optimal experiments as it allows to account for the uncertainty reduction obtained by transferring interventional data.

\section{Experiments}

\paragraph{Implementation details:} For all experiments we assume Gaussian distributions for the integrating measures and the conditional distributions in the \DAG{s} and optimise the parameters via maximum likelihood. We compute the integrals in \eqs (4)--(5) via Monte-Carlo integration with $1000$ samples. Finally, we fix the variance in the likelihood of \eq (3) and fix the kernel hyper-parameters for both the \rbf and causal kernel to standard values ($l=1$, $\sigma^2_f=1$). More works need to be done to optimise these settings potentially leading to improved performances. 

\subsection{$\DAG1$}

\paragraph{\textit{Do}-calculus derivations }
For $\DAG1$ (\fig 1 (a)) we have $\I = \{Z\}$ and  $\C = \emptyset$. The base function is thus given by $\basefunction = \expectation{}{Y|\DO{Z}{z}}$. In this section we give the expressions for the functions in $\T$ and show each of them can be written as a transformation of $\basefunction$ with the corresponding integrating measure.  Notice that in this case $\basefunction \in \T$.

\begin{align*}
\expectation{}{Y|\DO{X}{x}} &= \int \expectation{}{Y|\DO{X}{x}, z}p(z|\DO{X}{x}) \dint z
\\&= \int \expectation{}{Y|\DO{X}{x}, \DO{Z}{z}}p(z|\DO{X}{x}) \dint z \;\;\;  \text{by} \;\;\; Y \indep Z|X \text{ in } \mathcal{G}_{\overline{B}\overline{X}\overline{Z}}  
\\&= \int \expectation{}{Y|\DO{Z}{z}}p(z|\DO{X}{x}) \dint z \;\;\;  \text{by} \;\;\; Y \indep X|Z \text{ in } \mathcal{G}_{\overline{XZ}}  
\\&= \int \basefunction(z) p(z|\DO{X}{x}) \dint z
\end{align*}
with $p(z|\DO{X}{x}) = p(z|X=x)$.

\begin{align*}
\expectation{}{Y|\DO{Z}{z}} &= \basefunction(z).
\end{align*}

\begin{align*}
\expectation{}{Y|\DO{X}{x}, \DO{Z}{z}} &= \expectation{}{Y|\DO{Z}{z}} = \basefunction(z) \\
&  \;\;\; \text{by} \;\;\; Y \indep X|Z \text{ in } \mathcal{G}_{\overline{XZ}}  
\end{align*}

\paragraph{\sem:} 
\begin{align*}
X &= \epsilon_{X} \\
Z &= \exp(-X) + \epsilon_{Z} \\
Y &= \cos(Z) - \exp(-Z/20) + \epsilon_Y\\
\end{align*}

We consider the following interventional domains:
\begin{itemize}
	\item $D(X) = [-5,5]$ 
	\item $D(Z) = [-5,20]$ 
\end{itemize}

\subsection{$\DAG2$}

\paragraph{\textit{Do}-calculus derivations}
For $\DAG2$ (\fig 1 (b)) we consider $\{A,C\}$ to be non-manipulative. We have $\I = \{D, E\}$ and  $\C = \{A, B\}$. The base function is thus given by $\basefunction = \expectation{}{Y|\DO{D}{d}, \DO{E}{e}, a, b}$. In this section we give the expressions for all the functions in $\T$ and show each of them can be written as a transformation of $\basefunction$ with the corresponding integrating measure.  

\paragraph{Intervention sets of size 1}


\begin{align*}
\expectation{}{Y|\DO{B}{b}} &= \int \expectation{}{Y|\DO{B}{b}, d, e, a} p(d,e,a|\DO{B}{b})\dint d \dint e \dint a \\
&= \int \expectation{}{Y|\DO{B}{b}, \DO{D}{d}, \DO{E}{e}, a} p(d,e,a|\DO{B}{b})\dint d \dint e \dint a   \\& \text{by} \;\;\; Y \indep D,E | B, A \text{ in } \mathcal{G}_{\overline{B} \underline{D}\underline{E}} \\
&=\int \expectation{}{Y| \DO{D}{d}, \DO{E}{e}, a} p(d,e,a|\DO{B}{b})\dint d \dint e \dint a    \;\;\;  \text{by} \;\;\; Y \indep B|D,E,A \text{ in } \mathcal{G}_{\overline{B}\overline{D}\overline{E}}  \\
&= \int  \expectation{}{Y| \DO{D}{d}, \DO{E}{e}, a, b'} p(b') p(d,e,a|\DO{B}{b}) \dint d \dint e \dint a \dint b'  \\
&= \int \basefunction(d,e,a,b') p(b') p(d,e,a|\DO{B}{b}) \dint d \dint e \dint a \dint b'
\end{align*}

with $p(b')p(d,e,a|\DO{B}{b}) = p(b')p(a)p(d|e,a,B=b)p(e|a, B=b)$.

%
%
\begin{align*}
\expectation{}{Y|\DO{D}{d}} &= \int \expectation{}{Y|\DO{D}{d}, e, a, b} p(a,b,e|\DO{D}{d})\dint a \dint b \dint e  \\
&= \int \expectation{}{Y|\DO{D}{d}, \DO{E}{e}, a, b} p(a,b,e|\DO{D}{d})\dint a \dint b \dint e \;\;\;  \text{by} \;\;\; Y \indep E |D,A,B \text{ in }   \mathcal{G}_{\overline{D}\underline{E}} \\
& = \int \basefunction(d,e,a,b)  p(a,b,e|\DO{D}{d})\dint a \dint b \dint e
\end{align*}

with $p(a,b,e|\DO{D}{d}) = p(a)p(b)p(e|a,b)$.

\begin{align*}
\expectation{}{Y|\DO{E}{e}}  &= \int \expectation{}{Y|\DO{E}{e},d,a,b}  p(d,a,b|\DO{E}{e}) \dint a \dint b \dint s \\
&= \int \expectation{}{Y|\DO{E}{e},\DO{D}{d},a,b}  p(d,a,b|\DO{E}{e}) \dint a \dint b \dint d   \;\;\;  \text{by}  \;\;\; Y \indep D |E,A,B  \text{ in }   \mathcal{G}_{\overline{E}\underline{D}}\\
& = \int \basefunction(d,e,a,b)  p(d,a,b|\DO{E}{e}) \dint a \dint b \dint d 
\end{align*}

with $p(d,a,b|\DO{E}{e}) = p(a)p(b)p(d|b)$.

\paragraph{Intervention sets of size 2}

\begin{align*}
\expectation{}{Y|\DO{B}{b}, \DO{D}{d}} &= \int \expectation{}{Y|\DO{B}{b}, \DO{D}{d},a,e}p(a,e|\DO{B}{b}, \DO{D}{d}) \dint a \dint e \\
&=\int \expectation{}{Y|\DO{B}{b}, \DO{D}{d},a,\DO{E}{e}}p(a,e|\DO{B}{b}, \DO{D}{d}) \dint a \dint e
\\& \text{by} \;\;\; Y \indep E | A,B,D \text{ in } \mathcal{G}_{\overline{BD} \underline{E}} \\
&=\int \expectation{}{Y|\DO{D}{d},\DO{E}{e},a}p(a,e|\DO{B}{b}, \DO{D}{d}) \dint a \dint e
\\& \text{by} \;\;\; Y \indep B | A,D,E \text{ in } \mathcal{G}_{\overline{BDE}} \\
&=\int \expectation{}{Y|\DO{D}{d},\DO{E}{e},a,b'}p(b')p(a,e|\DO{B}{b}, \DO{D}{d}) \dint a \dint b' \dint e
\end{align*}
with $p(b')p(a,e|\DO{B}{b}, \DO{D}{d}) = p(b')p(a)p(e|a, B=b)$.

\begin{align*}
\expectation{}{Y|\DO{B}{b}, \DO{E}{e}}  &= \int \expectation{}{Y|\DO{B}{b}, \DO{E}{e},a,d}   p(a,d|\DO{B}{b}, \DO{E}{e}) \dint a \dint d \\
&= \int \expectation{}{Y|\DO{B}{b}, \DO{E}{e},a,\DO{D}{d}}   p(a,d|\DO{B}{b}, \DO{E}{e}) \dint a \dint d 
\\& \text{by} \;\;\; Y \indep D | A,B,E \text{ in } \mathcal{G}_{\overline{BE} \underline{D}} \\
&= \int \expectation{}{Y|\DO{D}{d}, \DO{E}{e},a}   p(a,d|\DO{B}{b}, \DO{E}{e}) \dint a \dint d 
\\& \text{by} \;\;\; Y \indep B | A,D,E \text{ in } \mathcal{G}_{\overline{BDE}} \\
&= \int \expectation{}{Y|\DO{D}{d}, \DO{E}{e},a,b'} p(b')  p(a,d|\DO{B}{b}, \DO{E}{e}) \dint a \dint b' \dint d \\
&= \int \basefunction(d,e,a,b')p(b')  p(a,d|\DO{B}{b}, \DO{E}{e}) \dint a \dint b' \dint d 
\end{align*}

with $p(b')  p(a,d|\DO{B}{b}, \DO{E}{e}) = p(b')p(a)p(d|B=b)$.

\begin{align*}
\expectation{}{Y|\DO{D}{d}, \DO{E}{e}} &= \int \expectation{}{Y|a,b,\DO{D}{d}, \DO{E}{e}} p(a,b |\DO{D}{d}, \DO{E}{e})\dint a \dint b \\
&= \int \basefunction(d,e,a,b) p(a,b |\DO{D}{d}, \DO{E}{e}) \dint a \dint b   
\end{align*}

with $p(a,b |\DO{D}{d}, \DO{E}{e}) = p(a)p(b)$.

\paragraph{Intervention sets of size 3}

\begin{align*}
\expectation{}{Y|\DO{B}{b}, \DO{D}{d},  \DO{E}{e}}  &= \expectation{}{Y| \DO{D}{d},  \DO{E}{e}} \\& \text{by} \;\;\; (Y \indep B|D, E \text{ in }    \mathcal{G}_{ \overline{D} \overline{E}\overline{B}})
\end{align*}

\paragraph{\sem:} 
\begin{align*}
U_1 &= \epsilon_{YA} \\
U_2 &= \epsilon_{YB} \\
A &= U_1 + \epsilon_A\\
B &=  U_2 + \epsilon_B\\
C &= \exp(-B) + \epsilon_C \\
D &= \exp(-C)/10. + \epsilon_D \\
E &= \text{cos}(A) + C/10 + \epsilon_E \\
Y &= \text{cos}(D) + \text{sin}(E) + U_1 + U_2  + \epsilon_y \\
\end{align*}

We consider the following interventional domains:
\begin{itemize}
	\item $D(B) = [-3,4]$ 
	\item $D(D) = [-3,3]$ 
	\item $D(E) = [-3,3]$ 
\end{itemize}

\subsection{$\DAG3$}

\paragraph{\textit{Do}-calculus derivations}
For $\DAG3$ (\fig 1 (c)) we consider $\{\text{age}, \text{\acro{bmi}}, \text{cancer}\}$ to be non-manipulative. We have $\I = \{\text{aspirin}, \text{statin}, \text{age}, \text{\acro{bmi}}, \text{cancer}\}$ and  $\C = \emptyset$. In this section we give the expressions for all the functions in $\T$ and show each of them can be written as a transformation of $\basefunction$ with the corresponding integrating measure.  

\begin{align*}
\expectation{}{Y|\DO{\text{aspirin}}{x}} &=  \idotsint f(\text{aspirin}, \text{statin}, \text{age}, \text{\acro{bmi}},\text{cancer}) \\
&  \;\;\;\;\;\;\;\;\; p( \text{statin}, \text{age}, \text{\acro{bmi}},\text{cancer}|\DO{\text{aspirin}}{x}) \text{d}\text{statin}\text{d}\text{age}\text{d}\text{\acro{bmi}}\text{d}\text{cancer}
\end{align*}
with $p( \text{statin}, \text{age}, \text{\acro{bmi}},\text{cancer}|\DO{\text{aspirin}}{x}) = p(\text{cancer}|\text{age}, \text{\acro{bmi}}, \text{aspirin}, \text{aspirin})p(\text{statin}|\text{age}, \text{\acro{bmi}})p(\text{\acro{bmi}}|\text{age})p(\text{age})$.

\begin{align*}
\expectation{}{Y|\DO{\text{statin}}{x}} &=  \idotsint f(\text{aspirin}, \text{statin}, \text{age}, \text{\acro{bmi}},\text{cancer}) \\
&  \;\;\;\;\;\;\;\;\; p( \text{aspirin}, \text{age}, \text{\acro{bmi}},\text{cancer}|\DO{\text{statin}}{x}) \text{d}\text{aspirin}\text{d}\text{age}\text{d}\text{\acro{bmi}}\text{d}\text{cancer}
\end{align*}
with $p( \text{aspirin}, \text{age}, \text{\acro{bmi}},\text{cancer}|\DO{\text{statin}}{x}) = p(\text{cancer}|\text{age}, \text{\acro{bmi}}, \text{aspirin}, \text{aspirin})p(\text{aspirin}|\text{age}, \text{\acro{bmi}})p(\text{\acro{bmi}}|\text{age})p(\text{age})$.

\begin{align*}
\expectation{}{Y|\DO{\text{aspirin}}{x}, \DO{\text{statin}}{z}} =  &=  \idotsint f(\text{aspirin}, \text{statin}, \text{age}, \text{\acro{bmi}},\text{cancer}) \\
&  \;\;\;\;\;\;\;\;\; p(\text{age}, \text{\acro{bmi}},\text{cancer}|\DO{\text{aspirin}}{x}, \DO{\text{statin}}{z}) \text{d}\text{age}\text{d}\text{\acro{bmi}}\text{d}\text{cancer}
\end{align*}
with $p(\text{age}, \text{\acro{bmi}},\text{cancer}|\DO{\text{aspirin}}{x}, \DO{\text{statin}}{z}) = p(\text{cancer}|\text{age}, \text{\acro{bmi}}, \text{aspirin}, \text{aspirin})p(\text{\acro{bmi}}|\text{age})p(\text{age})$.

\paragraph{\sem:} 

\begin{align*}
\text{age} &= \mathcal{U}(55,75) \\
\text{bmi} &= \mathcal{N}(27.0 - 0.01 \times \text{age} , 0.7) \\
\text{aspirin} &= \sigma(-8.0 + 0.10\times \text{age}  + 0.03 \times \text{bmi} ) \\
\text{statin} &= \sigma(-13.0 + 0.10\times \text{age}  + 0.20\times \text{bmi} )\\
\text{cancer} &=  \sigma(2.2 - 0.05\times \text{age}  + 0.01 \times \text{bmi}  - 0.04 \times \text{statin} + 0.02\times \text{aspirin} ) \\
Y &=  \mathcal{N}(6.8 + 0.04 \times \text{age}  - 0.15 \times \text{bmi}  - 0.60 \times \text{statin} + 0.55 \times \text{aspirin}  + 1.00 \times \text{cancer} , 0.4) \\
\end{align*}

We consider the following interventional domains:
\begin{itemize}
	\item $D(\text{aspirin}) = [0,1]$ 
	\item $D(\text{statin}) = [0,1]$ 
\end{itemize}

\subsection{Additional experimental results}
Here we give additional experimental results for both the synthetic examples and the health-care application. \tbl \ref*{table500} gives the fitting performances, across intervention functions and replicates, when $N=500$. 

\begin{table}[h]
		\caption{\acro{rmse} with $N=500$}
		\centering
		\begin{tabular}{@{\extracolsep{4pt}}@{\kern\tabcolsep}lcccccccccccccccc}
			\toprule
			&$\our^+$ & \our & $\gptext^+$ &  \gptext &\textit{do}-calculus \\
			\cmidrule(lr){2-6}
			\multirow{2}{*}{$\DAG1$}  &  \textbf{0.48}  &  0.57 & 0.60 & 0.77 &  0.55 \\
			& (0.07) &(0.08) & (0.15) & (0.27) & -  \\
			\multirow{2}{*}{$\DAG3$}     &  0.50 &  \textbf{0.42} & 0.58 & 1.26 & 2.87 \\
			&(0.11)&(0.13)  & (0.10) & (0.11)  & -  \\
			\multirow{2}{*}{$\DAG4$}     & \textbf{0.09} & 0.44 & 0.54 & 0.89  & 0.22 \\
			&(0.05)&(0.12)  & (0.08) & (0.23)  & -  \\
			\bottomrule
		\end{tabular} \label{table500}
	\end{table}

%
%

\bibliographystyle{apalike}
\bibliography{references}

%% file: preamble.tex
\usepackage[utf8]{inputenc} 
\usepackage[T1]{fontenc}    
\usepackage{hyperref}       
\usepackage{url}            
\usepackage{amsfonts}       
\usepackage{nicefrac}       
\usepackage{microtype}      
\usepackage{amsfonts}
\usepackage{amsmath}
\usepackage{amssymb}
\usepackage{makecell}
\usepackage{caption}
\usepackage{subcaption}
\usepackage{amsthm}
\usepackage[titlenumbered, ruled]{algorithm2e}
\usepackage{stackengine,mathtools}
\usepackage{enumitem}
\PassOptionsToPackage{numbers, compress}{natbib}

\usepackage{tikz}
\usetikzlibrary{fit,positioning}
\usetikzlibrary{bayesnet}
\usetikzlibrary{shapes,snakes}
\usetikzlibrary{arrows,patterns}
\usetikzlibrary{shapes,decorations,arrows,calc,arrows.meta,fit,positioning}
\tikzset{
	bidirected/.style={Latex-Latex,dashed}
}
\usetikzlibrary{calc}

\let\emptyset\varnothing

\theoremstyle{definition}
\newtheorem{definition}{Definition}[section]
\newtheorem{theorem}{Theorem}[section]
\newtheorem{corollary}{Corollary}[section]

\setlist[itemize]{leftmargin=10pt}

\usepackage{booktabs}
\usepackage{multirow}
\usepackage{wrapfig}
\usepackage{cancel}

\usepackage[preprint]{neurips_2020}

%% file: notation.tex
\newcommand{\mat}[1]{\mathbf{#1}}
\renewcommand{\vec}[1]{ \mathbf{#1} } 

\newcommand{\gp}{\mathcal{GP}}

\newcommand{\dataset}{\mathcal{D}}
\newcommand{\x}{\vec{x}}

\newcommand{\Y}{\mat{Y}}

\newcommand{\expectation}[2]{ \mathbb{E}_{#1}{\left[#2\right]} }

\newcommand{\DO}[2]{\text{do}\left(#1 = #2\right)}
\newcommand{\imeasure}[3]{\pi_{#1}\left(#2, #3\right)}

\newcommand{\graph}{\mathcal{G}}

\newcommand{\I}{\mat{I}}
\newcommand{\ivalue}{\mat{v}}

\newcommand{\C}{\mat{C}}

\newcommand{\Z}{\mat{Z}}
\newcommand{\X}{\mat{X}}
\newcommand{\T}{\mat{T}}

\newcommand{\Xsbase}{\I}

\newcommand{\cvalue}{\mat{c}}

\newcommand{\z}{\mat{z}}
\newcommand{\bvalue}{\mat{b}}

\newcommand{\partsX}{\mathcal{P}(\X)}

\newcommand{\basefunction}{f}

%% file: macros.tex
\newcommand{\eg}{e.g.\xspace}

\newcommand{\eq}{Eq.\xspace}
\newcommand{\eqs}{Eqs.\xspace}
\newcommand{\fig}{Fig.\xspace}

\newcommand{\tbl}{Tab.\xspace}

\newcommand{\acro}[1]{\textsc{#1}\xspace}
\newcommand{\gptext}{\acro{gp}}

\newcommand{\rbf}{\acro{rbf}}

\newcommand{\rmse}{\acro{rmse}}

\newcommand{\al}{\acro{al}}
\newcommand{\bo}{\acro{bo}}

\newcommand{\cbo}{\acro{cbo}}

\newcommand{\DAG}{\acro{dag}}

\newcommand{\sem}{\acro{scm}}

\newcommand{\psa}{\acro{psa}}

\newcommand{\our}{\acro{dag-gp}}

\newcommand{\mi}{\acro{mi}}

\newcommand{\parents}{\text{Pa}}

%% file: abstract.tex
\begin{abstract}
This paper studies the problem of learning the correlation structure of a set of intervention functions defined on the directed acyclic graph (\DAG) of a causal model. 
This is useful when we are interested in jointly learning the causal effects of interventions on different subsets of variables in a \DAG, which is common in field such as healthcare or operations research. 
We propose the first multi-task causal Gaussian process (\gptext) model, which we call \our, that allows for information sharing \emph{across} continuous interventions and \emph{across} experiments on different variables. \our accommodates different assumptions in terms of data availability and captures the correlation between functions lying in input spaces of different dimensionality via a well-defined integral operator. We give theoretical results detailing \emph{when} and \emph{how} the \our model can be formulated depending on the \DAG. We test both the quality of its predictions and its calibrated uncertainties. Compared to single-task models, \our achieves the best fitting performance in a variety of real and synthetic settings. In addition, it helps to select optimal interventions faster than competing approaches when used within sequential decision making frameworks, like active learning or Bayesian optimization.
\end{abstract}

%% file: intro.tex
\section{Introduction}
Solving decision making problems in a variety of domains such as healthcare, systems biology or operations research, requires experimentation. By performing interventions one can understand how a system behaves when an action is taken and thus infer the cause-effect relationships of a phenomenon.
For instance, in healthcare, drugs are tested in randomized clinical trials before commercialization. Biologists might want to understand how genes interact in a cell once one of them is knockout. 
Finally, engineers investigate the impact of design changes on complex physical systems by conducting experiments on digital twins \cite{ye2019digital}. Experiments in these scenarios are usually expensive, time-consuming, and, especially for field experiments, they may present ethical issues. Therefore, researchers generally have to trade-off cost, time, and other practical considerations to decide which experiments to conduct, if any, to learn about the system behaviour. 

Consider the causal graph in \fig \ref{fig:fig_crop} which describes how crop yield $Y$ is affected by soil fumigants $X$ and the level of eel-worm population at different times $\Z = \{Z_1, Z_2, Z_3\}$  \citep{cochran1957experimental, pearl1995causal}. By performing a set of experiments, the investigator aims at learning the \textit{intervention functions} relating the expected crop yield to each possible intervention set and level. Na\"ively, one could achieve that by modelling each intervention function separately. However, this approach would disregard the correlation structure existing across experimental outputs and would increase the computational complexity of the problem.
Indeed, the intervention functions are correlated and each experiment carries information about the yield we would obtain by performing alternative interventions in the graph. 
For instance, observing the yield when running an experiment on the \textit{intervention set} $\{X,Z_1\}$ and setting the value to the \textit{intervention value} $\{x,z_1\}$, provides information about the yield we would get from intervening only on $X$ or on $\{X, Z_1, Z_2, Z_3\}$. 
This paper studies how to jointly model such intervention functions so as to transfer knowledge across different experimental setups and integrate observational and interventional data. The model proposed here enables proper uncertainty quantification of the causal effects thus allowing to define optimal experimental design strategies.
\input{fig_crop}

%% file: fig_crop.tex
\begin{wrapfigure}{r}{0.3\linewidth}
\centering
\vspace{0.5cm}
\includegraphics[width=1.0\linewidth]{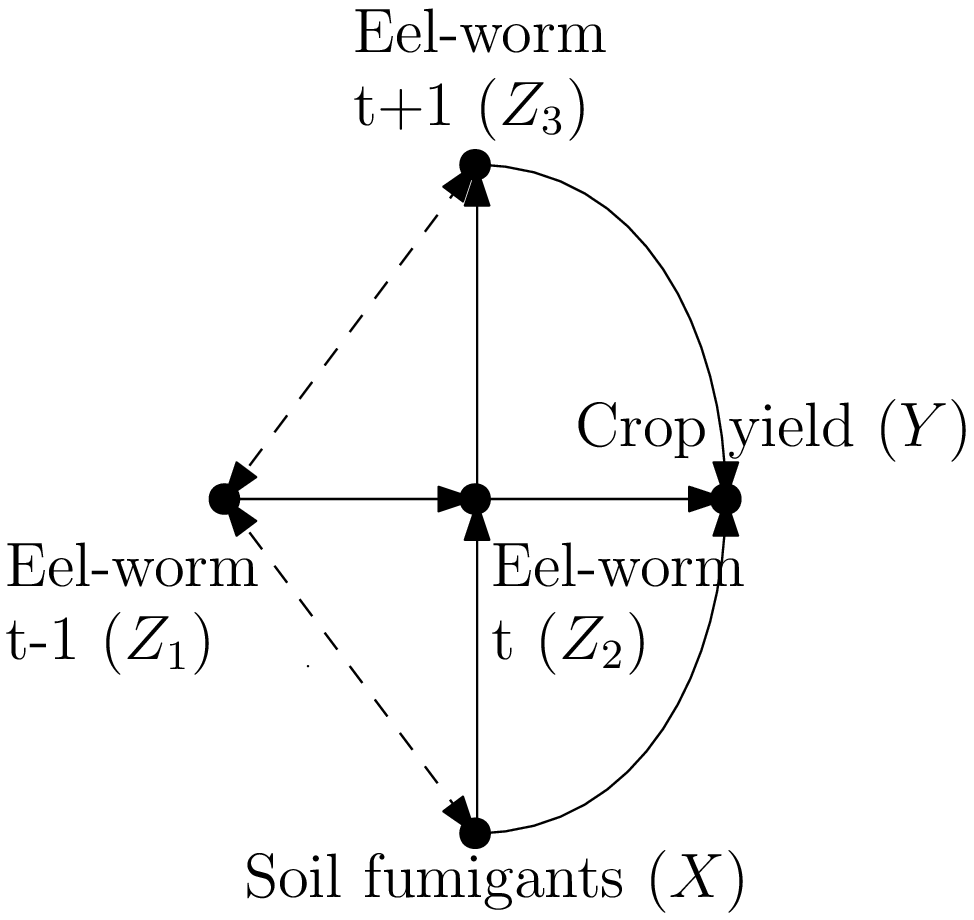} 
\caption{\DAG for the crop yield. Nodes denote variables, arrows represent causal effects and dashed edges indicate unobserved confounders.}
\label{fig:fig_crop}
\vspace{-1.3cm}
\end{wrapfigure}

%% file: motivation.tex
\subsection{Motivation and Contributions}
The framework proposed in this work combines causal inference with multi-task learning via Gaussian processes (\gptext, \cite{rasmussen2003gaussian}). Probabilistic causal models are commonly used in disciplines where explicit experimentation may be difficult and the \textit{do}-calculus \cite{pearl1995causal} allows to predict the effect of an intervention without performing the experiment. In \textit{do}-calculus, different intervention functions are modelled individually and there is no information shared across experiments. 
Modelling the correlation across experiments is crucial especially when the number of observational data points is limited and experiments on some variables cannot be performed. 
Multi-task \gptext methods have been extensively used to model non-trivial correlations between outputs \cite{alvarez2012kernels}. However, to the best of our knowledge, this is the first work focusing on intervention functions, possibly of different dimensionality, defined on a causal graph. Particularly, we make the following contributions:

\begin{itemize}
	\item We give theoretical results detailing \emph{when} and \emph{how} a causal multi-task model for the experimental outputs can be developed depending on the topology of the \DAG of a causal model. 
	\item Exploiting our theoretical results, we develop a joint probabilistic model for all intervention functions, henceforth named \our, which flexibly accommodates different assumptions in terms of data availability -- both observational and interventional.
	\item We demonstrate how \our achieves the best fitting performance in a variety of experimental settings while enabling proper uncertainty quantification and thus optimal decision making when used within Active Learning (\al) and Bayesian Optimization (\bo).
\end{itemize}

%% file: related.tex
\subsection{Related work} \label{sec:related}
While there exists an extensive literature on multi-task learning with \gptext{s} \cite{bonilla2008multi, alvarez2012kernels} and causality \citep{pearl2000causality, guo2018survey}, the literature on causal multi-task learning is very limited. 
The majority of the studies have focused on domain adaptation problems \cite{rojas2018invariant, magliacane2018domain, zhang2013domain} where data for a source domain is given, and the task is to predict the distribution of a target variable in a target domain. Several works \cite{ pearl2011transportability, bareinboim2012causal, bareinboim2013meta, bareinboim2014transportability} have studied the problem of transferring the causal effects of a given variable \emph{across} environments and have identified transportability conditions under which this is possible.
Closer to our work, \cite{alaa2017bayesian} have developed a linear coregionalization model for learning the individual treatment effects via observational data. While \cite{alaa2017bayesian} is the first paper conceptualizing causal inference as a multi-task learning problem, its focus is on modelling the correlation across intervention levels for a single intervention function corresponding to a dichotomous intervention variable. 

Differently from these previous works, this paper focuses on transfer \emph{within} a single environment, \emph{across} experiments and \emph{across} intervention levels. The set of functions we wish to learn have continuous input spaces of different dimensionality. Therefore, capturing their correlation requires placing a probabilistic model over the inputs which enables mapping between input spaces. The \DAG, which we assumed to be known and is not available in standard multi-task settings, allows us to define such a model. Therefore, \textit{existing multi-output \gptext models are not applicable to our problem}. 

Our work is also related to the literature on causal decision making. Studies in this field have focused on multi-armed bandit problems \cite{bareinboim2015bandits,lattimore2016causal, lu2018deconfounding, lee2018structural} and reinforcement learning \cite{buesing2018woulda, foerster2018counterfactual} settings where arms or actions correspond to interventions on a \DAG. More recently, \cite{cbo} proposed a Causal Bayesian Optimization (\cbo) framework solving the problem of finding an optimal intervention in a \DAG by modelling the intervention functions with \gptext{s}. In \cbo each function is modelled independently and their correlation is not accounted for when exploring the intervention space. This paper overcomes this limitation by introducing a multi-task model for experimental outputs. 
Finally, in the causal literature there has been a growing interest for experimental design algorithms to learn causal graphs \cite{he2008active, hauser2014two, greenewald2019sample} or the observational distributions in a graph \cite{rubensteinALsem}. Here we use our multi-task model within an \al framework so as to efficiently learn the experimental outputs in a causal graph. 

%% file: background_problem.tex
\section{Background and Problem setup}
\input{sem}

\input{problem_def}

%% file: sem.tex
Consider a probabilistic structural causal model (\sem)  \citep{pearl2000causality} consisting of a directed acyclic graph $\graph$ (\DAG) and a four-tuple  $\langle \mat{U}, \mat{V}, F, P(\mat{U})\rangle$, where $\mat{U}$ is a set of independent \emph{exogenous} background variables distributed according to the probability distribution $P(\mat{U})$, $\mat{V}$ is a set of observed \emph{endogenous} variables and $F = \{f_1, \dots, f_{|\mat{V}|}\}$ is a set of functions such that $v_i = f_i(\parents_i, u_i)$ with $\parents_i = \parents(V_i)$ denoting the parents of $V_i$. $\graph$ encodes our knowledge of the existing causal mechanisms among $\mat{V}$.
 Within $\mat{V}$, we distinguish between two different types of variables: treatment variables $\mat{X}$ that can be manipulated and set to specific values\footnote{This setting can be extended to include non-manipulative variables. See  \cite{lee2019structural} for a definition of such nodes.} and output variables $\mat{Y}$ that represent the agent's outcomes of interest. Given $\graph$, we denote the \emph{interventional distribution} for two disjoint sets in $\mat{V}$, say $\mat{X}$ and $\mat{Y}$, as $P(\mat{Y}| \DO{\mat{X}}{\x})$. This is the distribution of $\mat{Y}$ obtained by intervening on $\mat{X}$ and fixing its value to $\x$ in the data generating mechanism, irrespective of the values of its parents. The interventional distribution differs from the \emph{observational distribution} which is denoted by $P(\mat{Y}| \mat{X = \x})$. 
Under some identifiability conditions \cite{galles2013identif}, \textit{do}-calculus allows to estimate interventional distributions and thus causal effects from observational distributions \citep{pearl1995causal}. 
In this paper, we assume the causal effect for $\X$ on $\Y$ to be identifiable $\forall \X \in \partsX$ with $\partsX$ denoting the power set of $\X$.

%% file: problem_def.tex
\subsection{Problem setup} \label{sec:problem_def}
Consider a \DAG $\graph$ and the related \sem. Define the set of intervention functions for $Y$ in $\graph$ as:
\begin{align}
\mat{T} = \{t_s(\x)\}_{s=1}^{|\partsX|}  \quad \quad t_s(\x) = \expectation{p(Y|\DO{\X_s}{\x})}{Y} = \expectation{}{Y|\DO{\X_s}{\x}} \text{.}
\end{align}
with $\X_s \in \partsX$ where $\partsX$ is the power set of $\X$ minus the empty set\footnote{We exclude the empty set as it corresponds to the observational distribution $t_{\emptyset}(\x) = \expectation{}{Y}$.} and $\x \in D(\X_s)$ where $D(\X_s) = \times_{X\in \X_s} D(X)$ with $D(X)$ denoting the \textit{interventional domain} of $X$.
Let $\dataset^O=\{\x_n, y_n\}_{n=1}^N$, with $\x_n \in \mathbb{R}^{|\X|}$ and $y_n \in \mathbb{R}$, be an observational dataset of size $N$ from this \sem. 
Consider an interventional dataset $\dataset^I = (\X^I, \Y^I)$ with $\X^I = \bigcup_s\{\x_{si}^I\}_{i=1}^{N^I_s}$ and $\Y^I = \bigcup_s\{y_{si}^I\}_{i=1}^{N^I_s}$ denoting the intervention levels and the function values observed from previously run experiments across sets in $\partsX$. $N^I_s$ represents the number of experimental outputs observed for the intervention set $\X_s$. Our goal is to define a joint prior distribution $p(\T)$ and compute the posterior $p(\T|\dataset^I)$ so as to make probabilistic predictions for $\T$ at some unobserved intervention sets and levels. 

%% file: methodology.tex
\section{Multi-task learning of intervention functions}

In this section we address the following question: \emph{can we develop a joint model for the functions $\T$ in a causal graph and thus transfer information across experiments? }

To answer this question we study the correlation among functions in $\T$ which varies with the topology of $\graph$. Inspired by previous works on latent force models \cite{alvarez2009latent}, we show how any  functions in $\T$ can be written as an integral transformation of some base function $\basefunction$, also defined starting from $\graph$, via some integral operator $L_s$ such that $t_s(\x) = L_s(\basefunction)(\x)\text{,} \;\forall \X_s \in \partsX$. We first characterize the latent structure among experimental outputs and provide an explicit expression for both $\basefunction$ and $L_s$ for each intervention set (\S \ref{sec:transfer_conditions}). Based on the properties of $\graph$, we clarify when this function exists. Exploiting these results, we detail a new model to learn $\T$ which we call the \our model (\S \ref{sec:model}). In \our we place a \gptext prior on $\basefunction$ and propagate our prior assumptions on the remaining part of the graph to analytically derive a joint distribution of the elements in $\T$. The resulting prior distribution incorporates the causal structure and enables the integration of observational and interventional data. 
 
\input{theory}
\input{dag_gp_model}

%% file: theory.tex
\subsection{Characterization of the latent structure in a \DAG} \label{sec:transfer_conditions}
Next results provide a theoretical foundation for the multi-task causal \gptext model introduced later. 
In particular, they characterize when $\basefunction$ and $L_s$ exist and how to compute them thus fully characterizing when transfer across experiments is possible. All proofs are given in the appendix.

\begin{definition}
Consider a \DAG $\graph$ where the treatment variables are denoted by $\X$. Let $\C$ be the set of variables directly confounded with $Y$, $\C^{N}$ be the set of variables in $\C$ that are not colliders\footnote{Variables in $\C$ causally influenced by $\X$ and $Y$.} and $\I$ be the set $\parents(Y)$. For each $\X_s \in \partsX$ we define the following sets:
\begin{itemize}
	\item $\Xsbase^{N}_s = \Xsbase \backslash (\X_s \cap \Xsbase)$ represents the set of variables in $\I$ not included in $\X_s$.
	\item $\C^I_s = \C^{N} \cap \X_s$ is the set of variables in $\C$ which are included in $\X_s$ and are not colliders.
	\item $\C^{N}_s = \C^{N} \backslash \C^I_s$ is the set of variables in $\C$ that are neither included in $\X_s$ nor colliders.
\end{itemize}
\end{definition}

In the following theorem $\ivalue^{N}_s$ gives the values for the variables in the set $\Xsbase^{N}_s$ while $\cvalue$ represents the values for the set $\C^{N}$ which are partition in $\cvalue^{N}_s$ and $\cvalue^I_s$ depending on the set $\X_s$ we are considering.

\begin{theorem}\label{theorem_transfer} \textbf{Causal operator.}
\itshape
Consider a causal graph $\graph$ and a related \sem where the output variable and the treatment variables are denoted by $Y$ and $\X$ respectively. Denote by $\C$ the set of variables in $\graph$ that are \textit{directly} confounded with $Y$ and let $\I$ be the set $\parents(Y)$. Assume that $\C$ does not include nodes that have both \textit{unconfounded} incoming and outcoming edges. It is possible to prove that, $\forall \X_s \in \partsX$, the intervention function $t_s(\x):D(\X_s) \rightarrow \mathbb{R}$ can be written as $t_s(\x) = L_s(\basefunction)(\x)$ where 
\begin{align} \label{eq:eq_theorem}
L_s(f)(\x) = \idotsint \pi_s(\x, (\ivalue^{N}_s, \cvalue)) \basefunction(\ivalue, \cvalue)  \text{d} \ivalue^{N}_s \text{d}\cvalue \text{,}
\end{align} 
with $\basefunction(\ivalue, \cvalue) =  \expectation{}{Y|\DO{\Xsbase}{\ivalue}, \C^{N}=\cvalue}$ representing a shared latent function and $\pi_s(\x, (\ivalue^{N}_s, \cvalue)) = p(\cvalue^I_s|\cvalue^{N}_s)p(\ivalue^{N}_s, \cvalue^{N}_s| \DO{\X_s}{\x})$ giving the integrating measure for the set $\X_s$. 
\end{theorem}
In the sequel we call $L_s(\basefunction)(\x)$ the \textit{causal operator}, $(\I \cup \C)$ the \emph{base set}, $\basefunction(\ivalue, \cvalue)$ the \emph{base function} and $\pi_s(\cdot,\cdot)$ the \emph{integrating measure} of the set $\X_s$. A simple limiting case arises when the \DAG does not include variables directly confounded with $Y$ or $\C$ only includes colliders. In this case $\C=\emptyset$ and the base function is included in $\T$. 
Theorem \ref{theorem_transfer} provides a mechanism to reconstruct all causal effects emerging from $\partsX$ using the base function as a ``driving force''. In particular, the integrating measures can be seen as Green's functions incorporating the \DAG structure \cite{alvarez2009latent}.
While it can be further generalized to select $\Xsbase$ to be different from $\parents(Y)$, this choice is particularly useful due to the following result. 

\begin{corollary} \textbf{Minimality of $\boldsymbol{\I}$.}
	\itshape
	The smallest set $\Xsbase$ for which \eq \eqref{eq:eq_theorem} holds is given by $\parents(Y)$.
\end{corollary}

The dimensionality of $\Xsbase$ when chosen as $\parents(Y)$ has properties that have been previously studied in the literature. In the context of optimization \cite{cbo}, it corresponds to the so-called causal intrinsic dimensionality, which refers to the effective dimensionality of the space in which a function is optimized when causal information is available.
The existence of $\basefunction$ depends on the properties of the nodes in $\C$ which also represents the smallest set for which \eq \eqref{eq:eq_theorem} holds (\S\ref{sec:additional_theorem} in the supplement).

\begin{theorem} \label{function_existence} \textbf{Existence of $\boldsymbol{\basefunction}$.}
	\itshape
	If $\C$ includes nodes that have both unconfounded incoming and outcoming edges the function $\basefunction$ does not exist.
\end{theorem}

When $\basefunction$ does not exist, full transfer across \emph{all} functions in $\T$ is not possible (\DAG{s} with red edges in \fig \ref{fig:examples_dags}). However, these results enable a model for partial transfer across a subset of $\T$ (\S \ref{sec:partial_transfer} supp.).

%% file: dag_gp_model.tex
\input{fig_tx}
\subsection{The \our model} \label{sec:model}
Next, we introduce the \DAG \gptext model based on the results from the previous section.

\textbf{Model Likelihood:} Let $\dataset^I = (\X^I, \Y^I)$ be the interventional dataset defined in Section \ref{sec:problem_def}. Denote by $\T^I$ the collection of intervention vector-valued functions computed at $\X^I$. Each entry $y_{si}^I$ in $\Y^I$, is assumed to be a noisy observation of the corresponding function $t_s$ at $\x_{i}^I$:
\begin{align}
y_{si}^I = t_s(\x_{i}^I) + \epsilon_{si},\,\, \text{for} \,\,\, s=1,\dots,|\mathcal{P}(\X)|\,\, \text{and}\,\, i=1,\dots, N_s^I, \label{eq:lik}
\end{align}
with $\epsilon_{si} \sim \mathcal{N}(0, \sigma^2)$. In compact form, the joint likelihood function is $p(\Y^I|\T^I, \sigma^2)= \mathcal{N}(\T^I, \sigma^2\mat{I})$. 

\textbf{Prior distribution on $\T$:}
To define a join prior on the set of intervention functions, $p(\T)$, we take the following steps. First, we follow \cite{cbo} to place a \emph{causal prior} on $\basefunction$, the base function of the \DAG. Second, we propagate this prior on $\basefunction$ through all elements in $\T$ via the causal operator in \eq \eqref{eq:eq_theorem}. 

\emph{Step 1, causal prior on the base function}: The key idea of the causal prior, already used in \cite{cbo}, is to use the observational dataset $\dataset^O$ and the \textit{do}-calculus to construct the prior mean and variance of a \gptext that is used to model an intervention function. Our aim is to compute such prior for the causal effect of the base set $\I \cup \C$ on $Y$. The causal prior has the benefit of carrying causal information but at the expense of requiring $\dataset^{O}$ to estimate the causal effect. Any sensible prior can be used in this step, so the availability of $\dataset^O$ is not strictly necessity. However, in this paper we stick to the causal prior since it provides an explicit way of combining experimental and observational data. 

For simplicity, in the sequel we use $\bvalue = (\ivalue, \cvalue)$ to denote in compact form the values of the variables in the base set $\I=\ivalue$ and $C=\cvalue$. Using \emph{do-calculus} we can compute  $\hat{\basefunction}(\bvalue) = \hat{\basefunction}(\ivalue, \cvalue) = \hat{\mathbb{E}}[Y|\DO{\I}{\ivalue}, \cvalue] $ and $ \hat{\sigma}(\bvalue) =\hat{\sigma}(\ivalue, \cvalue) =\hat{\mathbb{V}}[Y|\DO{\I}{\ivalue}, \cvalue]^{1/2}$ where $\hat{\mathbb{V}}$ and $\hat{\mathbb{E}}$ represent the variance and expectation of the causal effects estimated from $\dataset^O$. The \emph{causal prior} $\basefunction(\bvalue) \sim \gp(m(\bvalue), K(\bvalue, \bvalue^\prime))$ is defined to have prior mean and variance given by $m(\bvalue)  = \hat{\basefunction}(\bvalue)$ and 
$K(\bvalue, \bvalue^\prime) =  k_{\rbf}(\bvalue, \bvalue^\prime) + \hat{\sigma}(\bvalue)\hat{\sigma}(\bvalue^\prime)$ respectively. The term $k_{\rbf}(\bvalue, \bvalue^\prime) :=\sigma^2_f \exp(-||\bvalue - \bvalue^\prime ||^2/2l^2)$ denotes the radial basis function (\rbf) kernel, which is added to provide more flexibility to the model.

\emph{Step 2, propagating the distribution to all elements in $\T$:} In Section \ref{sec:transfer_conditions} we showed how, $\forall \X_s \in \partsX$, $t_s(\x) = L_s(f)(\x)$ with $\basefunction$ given by the intervention function defined in Theorem \ref*{theorem_transfer}. By linearity of the causal operator, placing a \gptext prior on $\basefunction$ induces a well-defined joint \gptext prior distribution on $\T$. In particular, for each $\X_s \in \partsX$, we have $t_{s}(\x) \sim \gp(m_s(\x), k_s(\x,\x'))$ with:
\begin{align}
m_s(\x) &= \idotsint m(\bvalue) \imeasure{s}{\x}{\bvalue_s}\text{d}\bvalue_s \label{eq:mean_t_s} \\
k_s(\x,\x^\prime) &= \idotsint K(\bvalue, \bvalue^\prime) \imeasure{s}{\x}{\bvalue_s} \imeasure{s}{\x^\prime}{\bvalue_s^\prime}  \text{d}\bvalue_s  \text{d}\bvalue_s^\prime \text{.} \label{eq:cov_t_s}
\end{align}
where $\bvalue_s = (\ivalue_s^{N}, \cvalue)$ is the subset of $\bvalue$ including only the $\ivalue$ values corresponding to the set $\Xsbase^{N}_s$.

Let $D$ be a finite set of inputs for the functions in $\T$, that is $D = \bigcup_s \{\x_{si}\}_{i=1}^M$. $\T$ computed in $D$ follows a multivariate Gaussian distribution that is $\T^D \sim \mathcal{N}(m_{\T}(D), K_{\T}(D, D))$ with $K_{\T}(D,D) = (K_{\T}(\x,\x'))_{\x \in D, \x' \in D}$ and $m_{\T}(D) = (m_{\T}(\x))_{\x \in D}$. In particular, for two generic data points $\x_{si}, \x_{s^\prime j} \in D$ with $s$ and $s'$ denoting two \emph{distinct} functions we have $m_{\T}(\x_{si}) = \expectation{}{t_s(\x_{i})} = m_s(\x_{i})$ and $K_{\T}(\x_{si},\x_{s^\prime j})=\text{Cov}[ t_s(\x_i),t_{s'}(\x_j)]$. 

When computing the covariance function across intervention sets and intervention levels we differentiate between two cases. When both $t_s$ and $t_{s'}$ are different from $\basefunction$, we have: 
\begin{align*} \label{eq:crosscovariance}
\text{Cov}[t_s(\x_i) ,t_{s'}(\x_j)] &= \idotsint K(\bvalue, \bvalue^\prime) \imeasure{s}{\x_i}{\bvalue_s} \imeasure{s^\prime}{\x_j}{\bvalue^\prime_{s^\prime}} \text{d}\bvalue_s \text{d} \bvalue^\prime_{s^\prime} \text{.}
\end{align*}
If one of the two functions equals $\basefunction$, this expression further reduces to:
\begin{align*}
\text{Cov}[ t_s(\x_i),t_{s'}(\x_j)] = \int  K(\bvalue, \bvalue^\prime) \imeasure{s^\prime}{\x_j}{\bvalue^\prime_{s^\prime}} \text{d}\bvalue^\prime_{s^\prime}.
\end{align*} 
Note that the integrating measures $\imeasure{s}{\cdot}{\cdot}$ and $\imeasure{s^\prime}{\cdot}{\cdot}$ allow to compute the covariance between points that are defined on spaces on possibly different dimensionality, \emph{a scenario that traditional multi-output \gptext models are unable to handle}. The prior $p(\T)$ enables to merge different data types and to account for the natural correlation structure among interventions defined by the topology of the \DAG. For this reason we call this formulation the \our model. The parameters in \eqs \eqref{eq:mean_t_s}--\eqref{eq:cov_t_s} can be computed in closed form only when $K(\bvalue, \bvalue^\prime)$ is an \rbf kernel and the integrating measures are assumed to be Gaussian distributions. In all other cases, one needs to resort to numerical approximations \eg Monte Carlo integration in order to compute the parameters of each $t_s(\x)$.

\textbf{Posterior distribution on $\T$:} The posterior distribution $p(\T^D|\dataset^I)$ can be derived analytically via standard \gptext updates. For any set $D$, $p(\T^D|\dataset^I)$ will be Gaussian with parameters $m_{\T|\dataset^I}(D) = m_{\T}(D) + K_{\T}(D,\X^I) [K_{\T}(\X^I,\X^I) + \sigma^2 \mat{I}] (\T^I-m_{\T}(\X^I))$ and $K_{\T|\dataset^I}(D,D) = K_{\T}(D,D)-K_{\T}(D,\X^I) [K_{\T}(\X^I,\X^I) + \sigma^2 \mat{I}] K_{\T}(\X^I,D)$. See \fig \ref{fig:toy_1} for an illustration of the \DAG-\gptext model.

%% file: fig_tx.tex
\begin{figure*}
	\centering
	\includegraphics[width=0.9\linewidth]{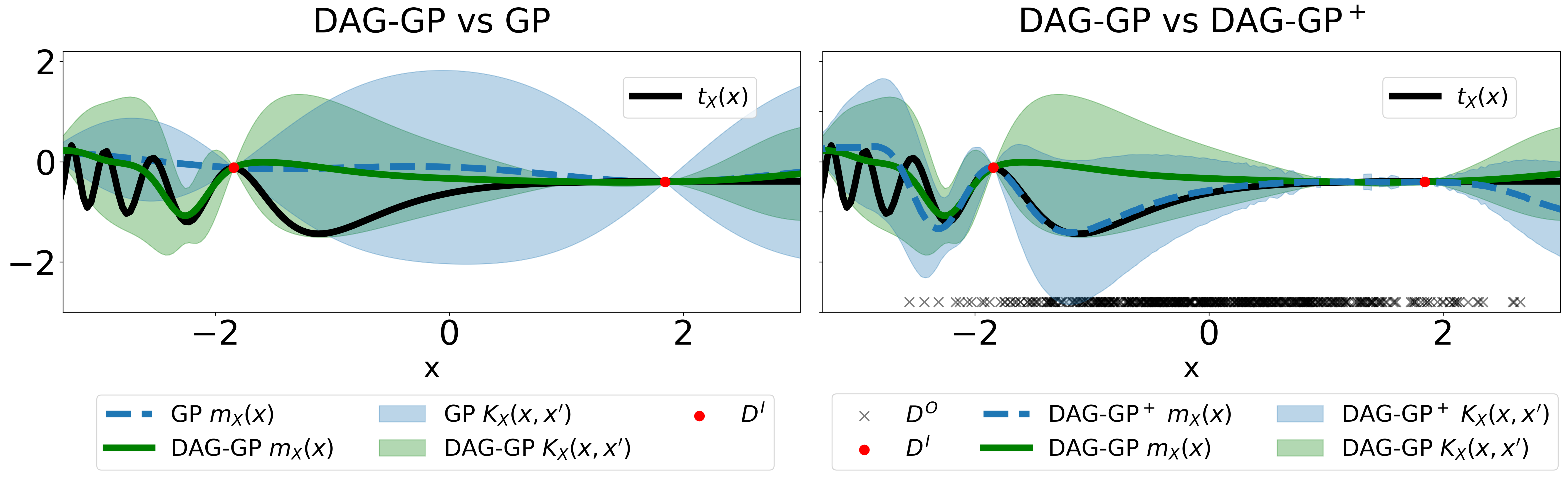} 
	\caption{Posterior mean and variance for $t_X(\x)$ in the \DAG of \fig \ref{fig:examples_dags} (a) (without the red edge). For both plots $m_X(\cdot)$ and $K_X(\cdot,\cdot)$ give the posterior mean and standard deviation respectively. 
	\emph{Left}: Comparison between the \our model and a single-task \gptext model (\gptext). \our captures the behaviour of $t_X(\x)$ in areas where $\dataset^I$ is not available (see area around $x=-2$) while reducing the uncertainty via transfer due to available data for $\z$ (see appendix). \emph{Right}: Comparison between \our with the causal prior ($\our^+$) and a standard prior with zero mean and \rbf kernel (\our). In addition to transfer, $\our^+$ captures the behaviour of $t_X(\x)$ in areas where $\dataset^O$ (black $\times$) is available (see region $[-2,0]$) while inflating the uncertainty in areas with no observational data.  }
	\label{fig:toy_1}
\end{figure*}

%% file: helicopter.tex
\input{fig_methods}
\section{A helicopter view} \label{sec:overview}
Different variations of the \our model can be considered depending on the availability of both observational $\dataset^O$ and interventional data $\dataset^I$ (\fig \ref{fig:map_methods}). Our goal here is not to be exhaustive, nor prescriptive, but to help to give some perspective. When $\dataset^I$ is not available \text{do}-calculus is the only way to learn $\T$, which in turns requires $\dataset^O$. When both data types are not available, learning $\T$ via a probabilistic model is not possible unless the causal effects can be transported from an alternative population.
In this case mechanistic models based on physical knowledge of the process under investigation are the only option. 
When $\dataset^I$  is available one can consider a single task or a multi-task model. If $\basefunction$ does not exist, a single \gptext model needs to be considered for each intervention function. Depending on the availability of $\dataset^O$, integrating observational data into the prior distribution (denoted by $\gptext^+$) or adopting a standard prior (denoted by $\gptext$) are the two alternatives. In both cases, the experimental information is not shared across functions and learning $\T$ requires intervening on all sets in $\partsX$. When instead $\basefunction$ exists, \our can be used to transfer interventional information and, depending on $\dataset^O$, also incorporating observational information a priori ($\our^+$).

%% file: fig_methods.tex
\begin{figure}
	\centering 
	\includegraphics[width=0.7\linewidth]{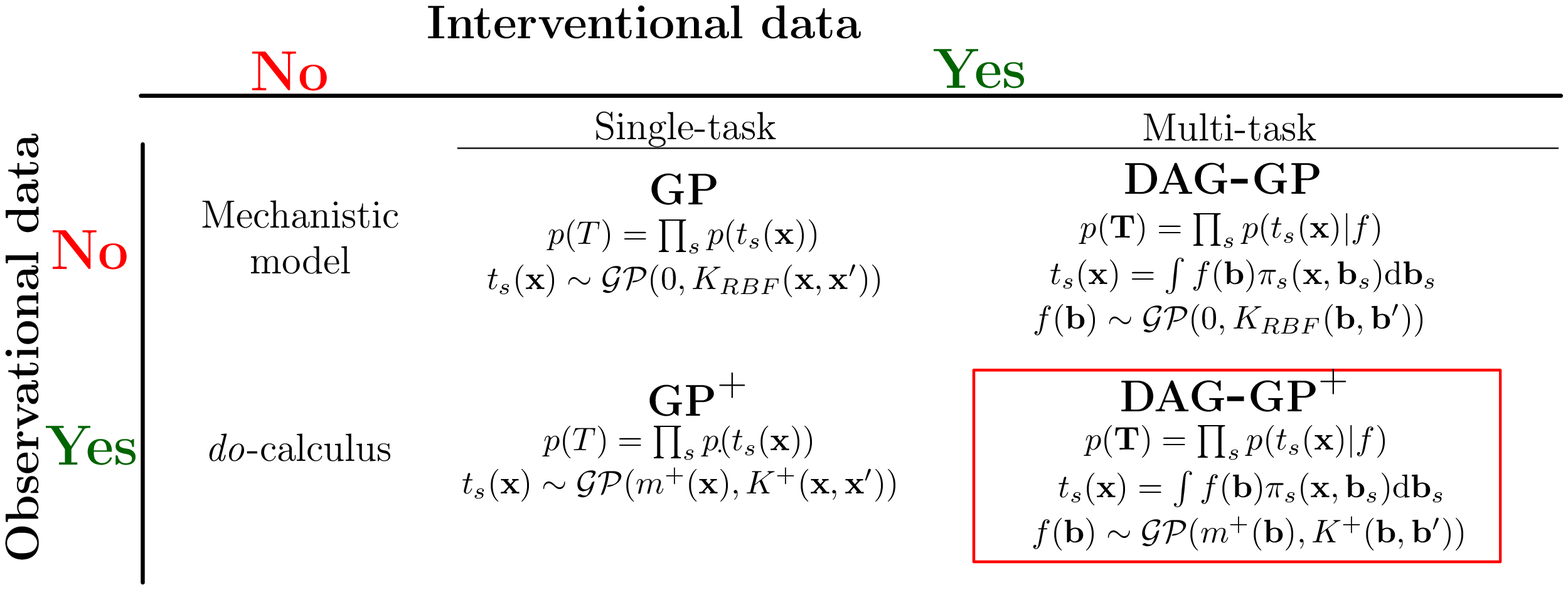}
	\caption{Models for learning the intervention functions $\T$ defined on a \DAG. The \textit{do}-calculus allows estimating $\T$ when only the observational data is available.  When the interventional data is also available, one can use a single-task model (denoted by \gptext) or a multi-task model (denoted by \our). When both data types are available one can combine them using the causal prior parameters represented by $m^+(\cdot)$ and $k^+(\cdot,\cdot)$. The resulting models are denoted by $\gptext^+$ and $\our^+$.}
	\label{fig:map_methods}
\end{figure}

%% file: experiments.tex
\section{Experiments}
This section evaluates the performance of the \our model on two synthetic settings and on a real world healthcare application (\fig \ref{fig:examples_dags}). We first learn $\T$  with fixed observational and interventional data (\S \ref{sec:fit_T}) and then use the \our model to solve active learning (\al) (\S \ref{sec:DM_AL}) and Bayesian Optimization (\bo) (\S \ref{sec:DM_CBO})\footnote{Code and data for all the experiments will be provided.}. Implementation details are given in the supplement.

\textbf{Baselines:} We run our algorithm both with ($\our^+$) and without ($\our$) causal prior and compare against the alternative models described in \fig \ref{fig:map_methods}.
Note that we do not compare against alternative multi-task \gptext models because, as mentioned in Section \ref{sec:related}, the models existing in the literature cannot deal with functions defined on different inputs spaces and thus can not be straightforwardly applied to our problem. 

\textbf{Performance measures:} We run all models with different initialisation of $\dataset^I$ and different sizes of $\dataset^O$. We report the root mean square error (\rmse) performances together with standard errors across replicates. For the \al experiments we show the \rmse evolution as the size of $\dataset^I$ increases. For the \bo experiments we report the convergence performances to the global optimum.

\input{fig_dags_in_exp}
\input{fit_functions}

\input{al_exp}

\input{cbo_exp}

%% file: fig_dags_in_exp.tex
\begin{figure}
	\centering
		\includegraphics[width=0.65\linewidth]{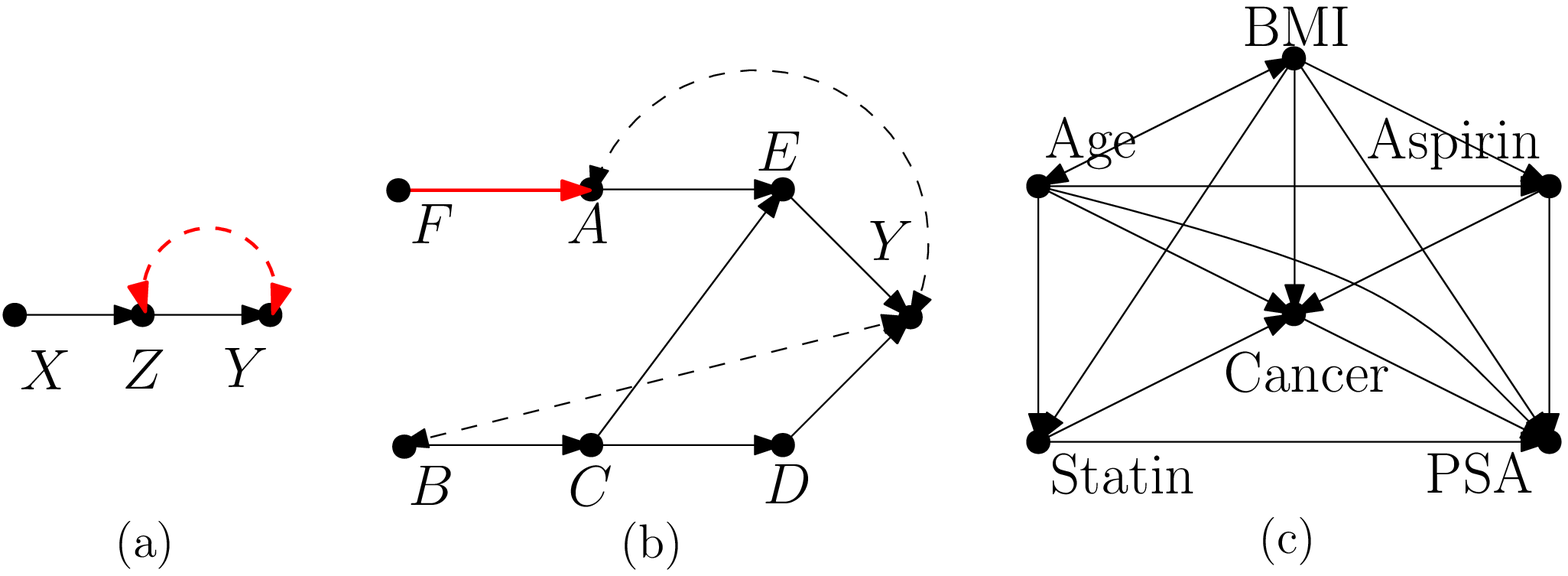} 
		\captionof{figure}{Examples of \DAG{s} (in black) for which $\basefunction$ exists and the \our model can be formulated. The red edges, if added, prevent the identification of $\basefunction$ making the transfer via \our not possible. }
		\label{fig:examples_dags}
\end{figure}

%% file: fit_functions.tex
\input{table_performances}

\subsection{Learning $\T$ from data}
\label{sec:fit_T}
\vspace{-0.2cm}
We test the algorithm on the \DAG{s} in \fig \ref{fig:examples_dags} and refer to them as (a) $\DAG1$, (b) $\DAG2$ and (c) $\DAG3$. $\DAG3$ is taken from \cite{Thompson} and \cite{ferro2015use} and is used to model the causal effect of statin drugs on the levels of prostate specific antigen (\psa). We consider the nodes $\{A,C\}$ in $\DAG2$ and $\{\text{age}, \acro{bmi}, \text{cancer}\}$ in $\DAG3$ to be non-manipulative. We set the size of $\dataset^I$ to $5\times |\T|$ for $\DAG1$ ($|\T|=2$), to $3\times |\T|$ for $\DAG2$ ($|\T|=6$) and to $|\T|$ for $\DAG3$ ($|\T|=3$). As expected, $\gptext^+$ outperforms $\gptext$ incorporating the information in $\dataset^O$ (\tbl \ref{table_100}). Interestingly, $\gptext^+$ also outperforms $\our$ in $\DAG3$ when $N=30$ and in $\DAG1$ when $N=100$. This depends on the effect that $\dataset^O$ has, through its size $N$ and its coverage of the interventional domains, on both the causal prior and the estimation of the integrating measures. 
Lower $N$ and coverage imply not only a less precise estimation of the \textit{do}-calculus but also a worse estimation of the integrating measures and thus a lower transfer of information. Higher $N$ and coverage imply more accurate estimation of the causal prior parameters and enhanced transfer of information across experiments. In addition, the way $\dataset^O$ affects the performance results it's specific to the $\DAG$ structure and to the distribution of the exogenous variables in the \sem.  
More importantly, \tbl \ref{table_100} shows how $\our^+$ consistently outperforms all competing methods by successfully integrating different data sources and transferring interventional information across functions in $\T$. Differently from competing methods, these results holds across different $N$ and $\dataset^I$ values making $\our^+$ a robust default choice for any application.

%% file: table_performances.tex
\begin{table}
	\caption{\rmse performances across 10 initializations of $\dataset^I$. See \fig \ref{fig:map_methods} for details on the compared methods. \textit{do} stands for the \textit{do}-calculus. $N$ is the size of $\dataset^O$. Standard errors in brackets.}
	\centering
\resizebox{\columnwidth}{!}{\begin{tabular}{@{\extracolsep{4pt}}@{\kern\tabcolsep}lcccccccccccccccc}
	\toprule
	&  \multicolumn{5}{c}{$N=30$} &\multicolumn{5}{c}{$N=100$} \\
	\cmidrule(lr){2-6} 
	\cmidrule(lr){7-11}
	&$\our^+$ & \our & $\gptext^+$ &  \gptext &\textit{do} &$\our^+$ & \our & $\gptext^+$ &  \gptext &\textit{do} \\
	\cmidrule(lr){2-6}  \cmidrule(lr){7-11}
	\multirow{2}{*}{$\DAG1$}  &   \textbf{0.46}  &  0.57 & 0.60 & 0.77 &  0.70 &\textbf{0.43}  &  0.57 & 0.45 & 0.77 &  0.52\\
	& (0.06) &(0.09) & (0.2) & (0.27) & - & (0.05) &(0.08) & (0.05) & (0.27) & -  \\
	\multirow{2}{*}{$\DAG2$}     &  \textbf{0.44} & 0.45 & 0.62 & 1.26 & 1.40 &  \textbf{0.36}  & 0.41 & 0.58 & 1.28 & 1.41\\
	&  (0.1) & (0.13)  & (0.10)  & (0.11) & - & (0.09) & (0.12) & (0.07) & (0.11) & -\\
	\multirow{2}{*}{$\DAG3$}     & \textbf{0.05} & 0.44 & 0.23 & 0.89 & 0.18 &  \textbf{0.06}  & 0.44 & 0.48 & 0.89 & 0.23\\
& (0.04) & (0.12)  & (0.03)  & (0.23) & -   & (0.04) & (0.12) & (0.06) & (0.23) & -\\
	\bottomrule
\end{tabular}}
\label{table_100}
\end{table}

%% file: al_exp.tex
\subsection{\our as surrogate model in Active Learning}\label{sec:DM_AL}
\vspace{-0.2cm}
\input{fig_al}
The goal of \al is to design a sequence of function evaluations to perform in order to learn a target function as quickly as possible.
We run \our within the \al algorithm proposed by \cite{krause2008} and select observations based on the Mutual Information (\mi) criteria extended to a multi-task setting (see \S \ref{sec:al_details} in the supplement for details). 
\fig \ref{fig:al_results} shows the \rmse performances as more interventional data are collected. Across different $N$ settings, $\our^+$ converges to the lowest \rmse performance faster then competing methods by collecting evaluations in areas where: (i) $\dataset^O$ does not provide information and (ii) the predictive variance is not reduced by the experimental information transferred from the other interventions.  
As mentioned before, $\dataset^O$ impacts on the causal prior parameters via the \textit{do}-calculus computations. When the latter are less precise, because of lower $N$ or lower coverage of the interventional domains, the model variances for $\our^+$ or $\gptext^+$ are inflated. Therefore, when $\our^+$ or $\gptext^+$ are used as surrogate models, the interventions are collected mainly in areas where $\dataset^O$ is not observed thus slowing down the exploration of the interventional domains and the convergence to the minimum \rmse (\fig \ref{fig:al_results} $\DAG2, N=100$).

%% file: fig_al.tex
\begin{figure} 
	\centering
	\includegraphics[width=0.325\linewidth]{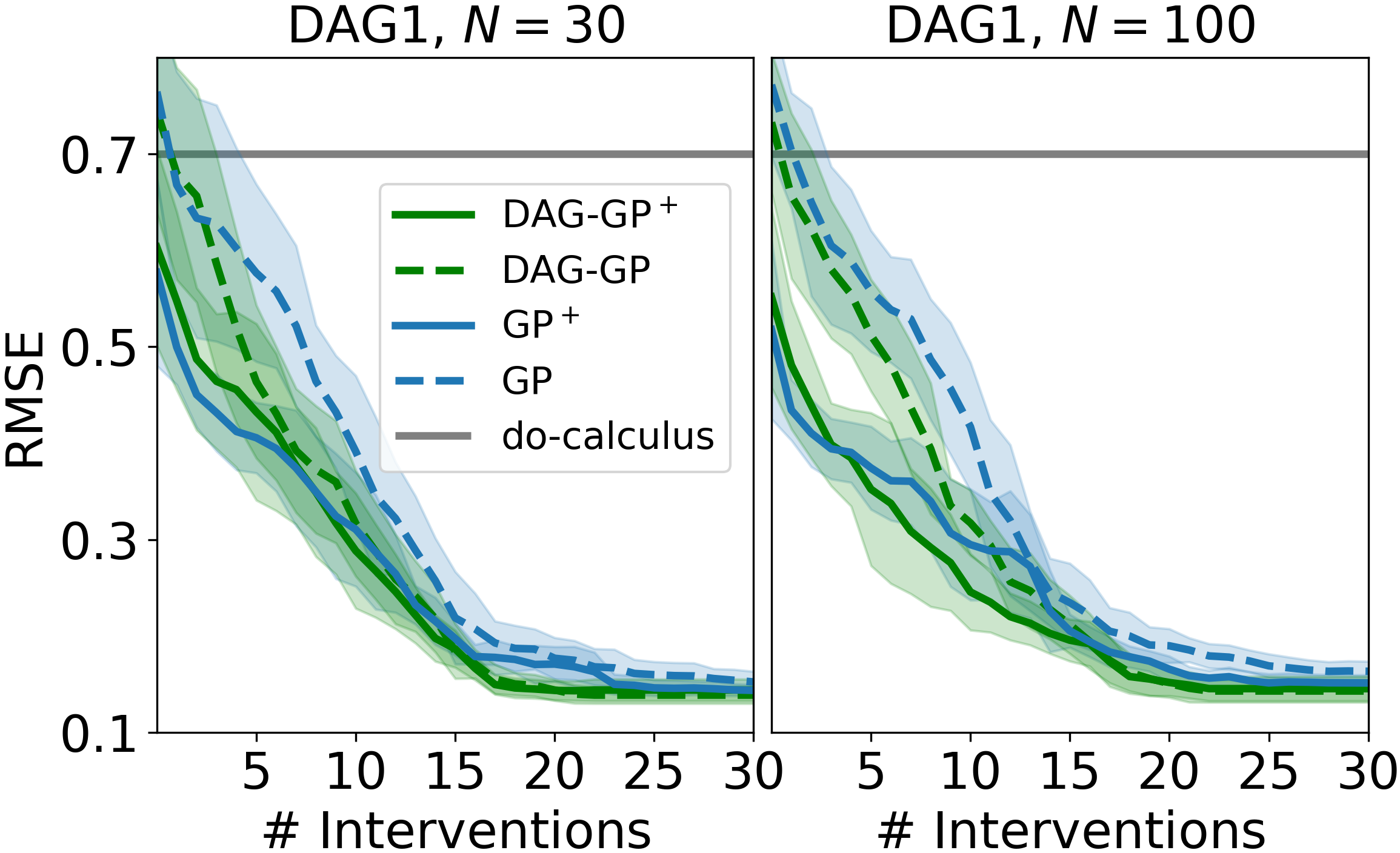} 
	\includegraphics[width=0.325\linewidth]{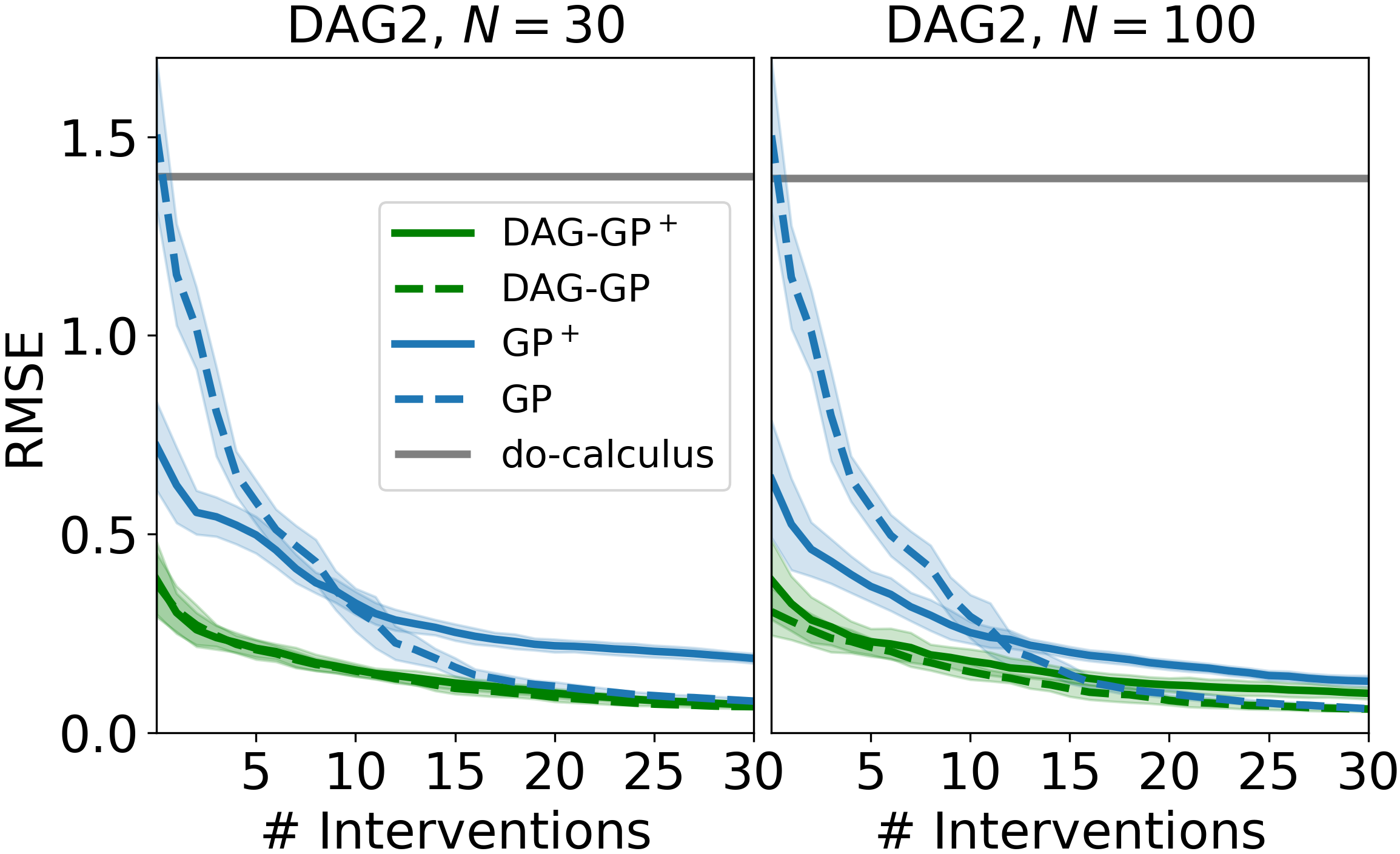} 
	\includegraphics[width=0.325\linewidth]{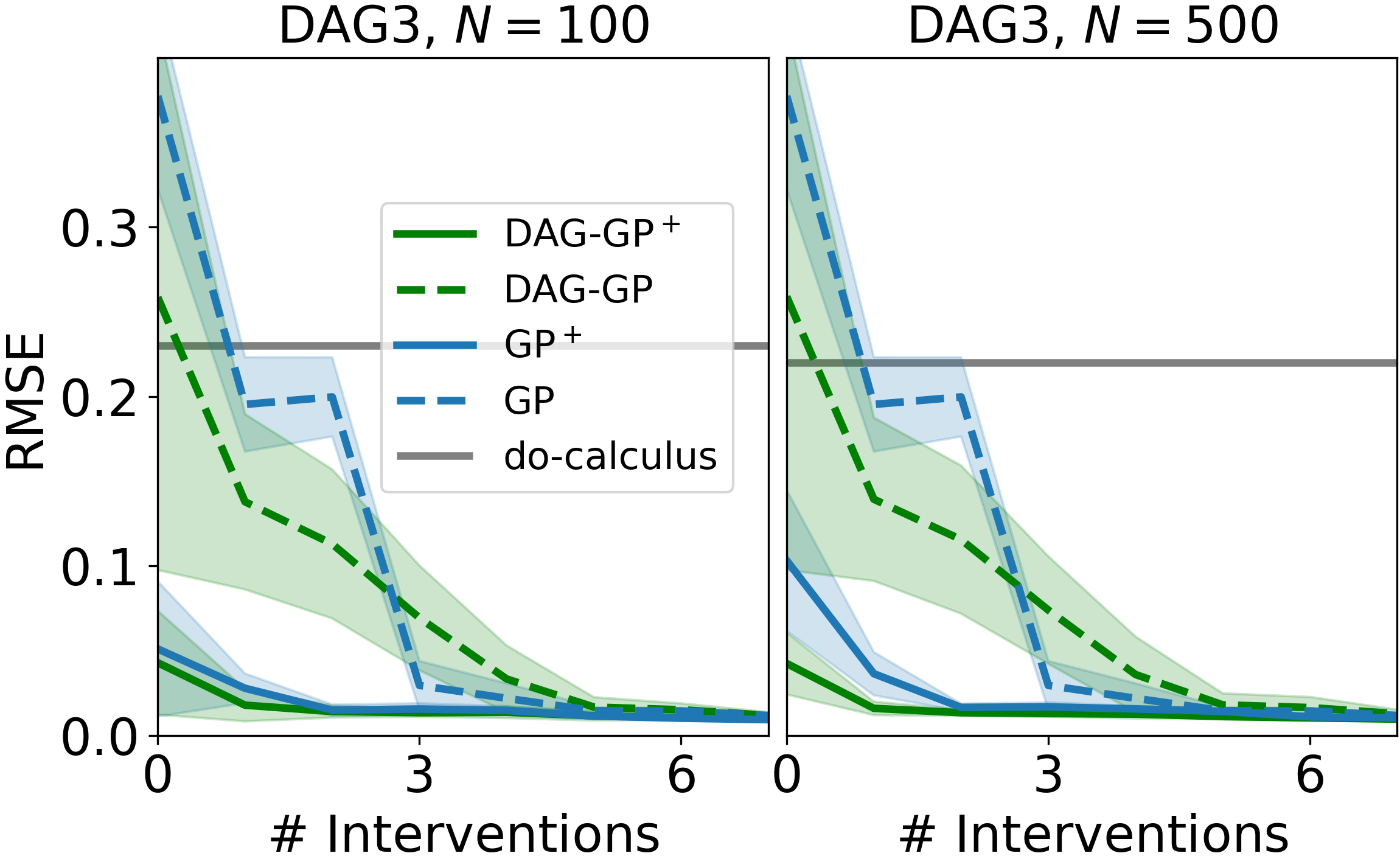} 
	\caption{\al results. Convergence of the \rmse performance across functions in $\T$ and across replicates as more experiments are collected. $\our^+$ gives our algorithm with the causal prior while $\our$ is our algorithm with a standard prior. $\#$ interventions is the number of experiments for each $\X_s$. Shaded areas give $\pm$ standard deviation.  See \fig \ref{fig:map_methods} for details on the compared methods.}
	\label{fig:al_results}
\end{figure}

%% file: cbo_exp.tex
\subsection{\our as surrogate model in Bayesian optimization}\label{sec:DM_CBO}
\input{fig_cbo}
\vspace{-0.2cm}
The goal of \bo is to optimize a function which is costly to evaluate and for which an explicit functional form is not available by making a series of function evaluations.
We use $\our$ within the \cbo algorithm proposed by \cite{cbo} (\fig \ref{fig:al_bo_complete} right plot) where a modified version of the expected improvement is used as an acquisition functions to explore a set of intervention functions.  
We compare $\our$ against the single-task models used in \cite{cbo}. We found \our to significantly speed up the convergence of \cbo to the global optimum both with and without the causal prior. 

%% file: fig_cbo.tex
 \begin{figure} 
	\centering
	\includegraphics[width=0.295\linewidth]{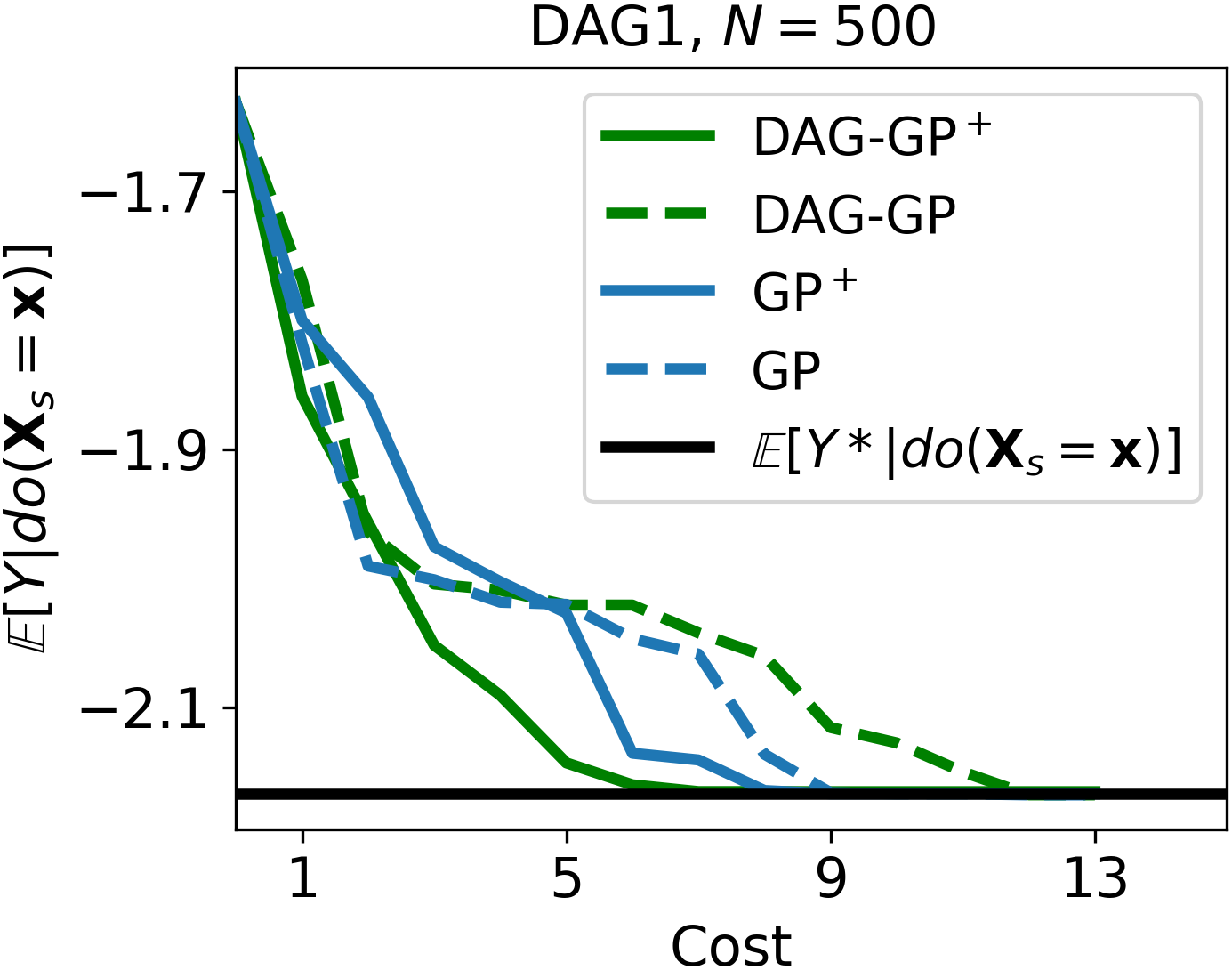}
	\includegraphics[width=0.3\linewidth]{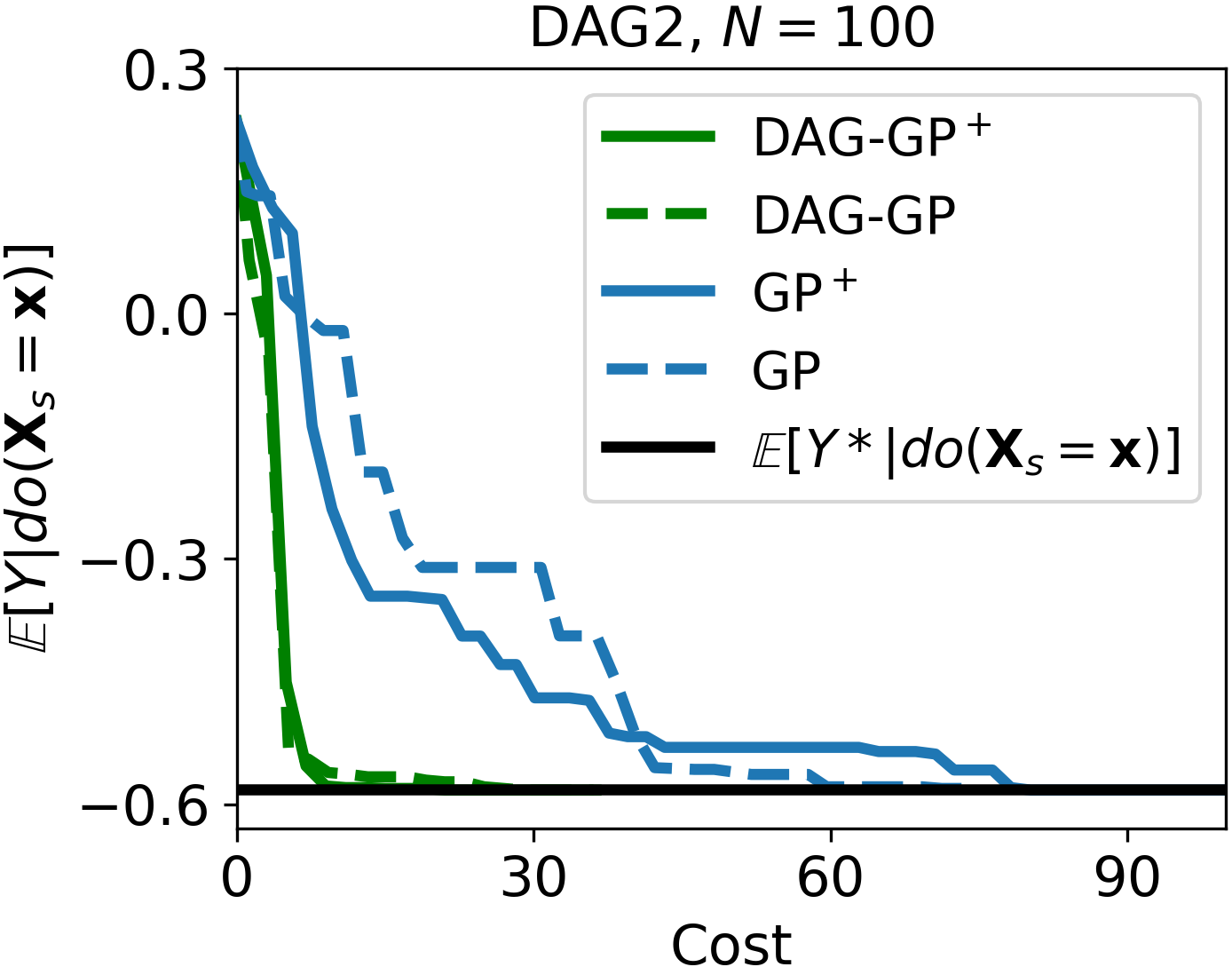}
	\includegraphics[width=0.3\linewidth]{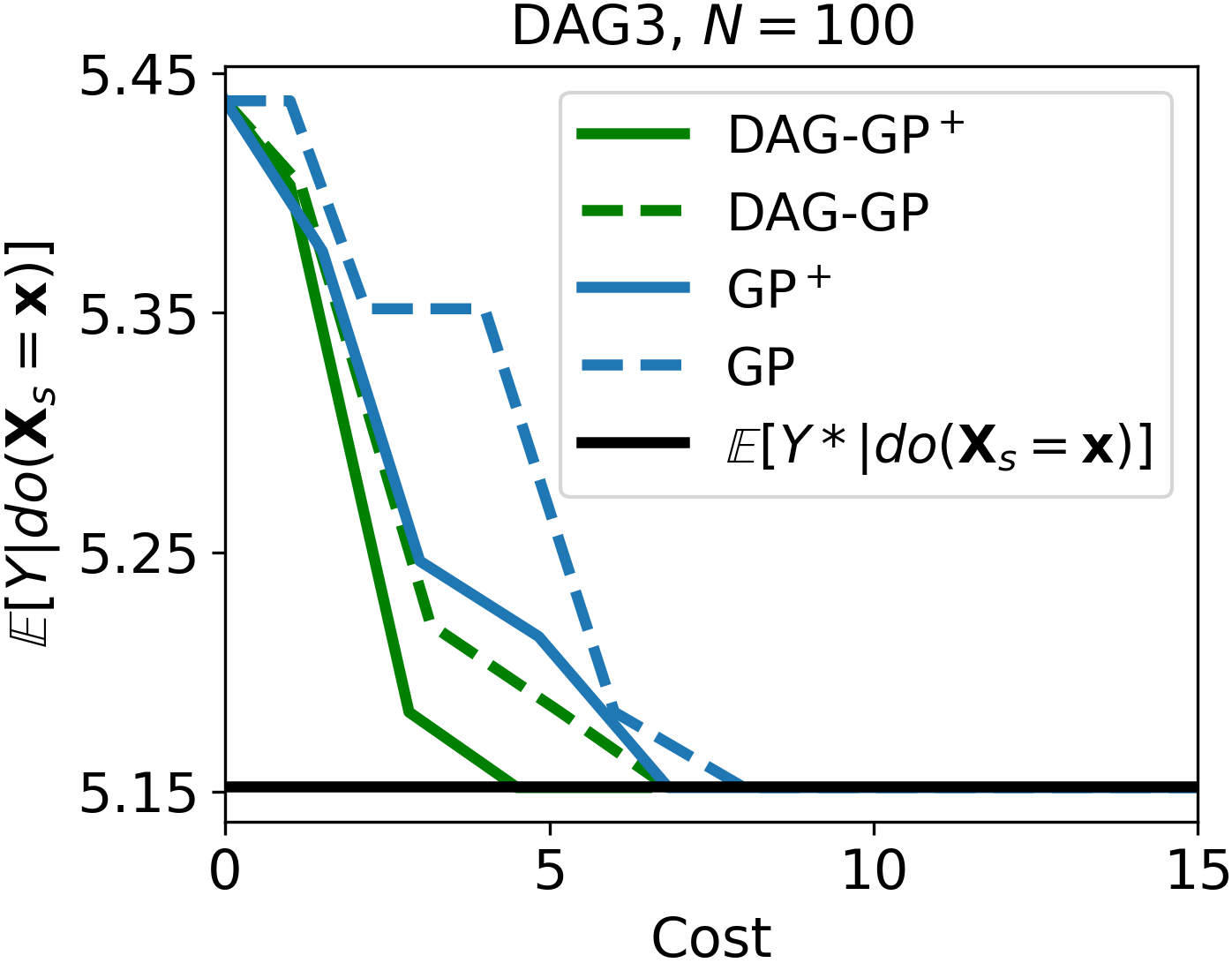}
	\caption{\bo results. Convergence of the \cbo algorithm to the global optimum ($\expectation{}{Y^\star | \DO{\X_s}{\x}}$) when our algorithm is used as a surrogate model with ($\our^+$) and without ($\our$) the causal prior. 
	See the supplement for standard deviations across replicates.}
	\label{fig:al_bo_complete}
\end{figure}

%% file: conclusions.tex
\section{Conclusions}
\vspace{-0.3cm}
This paper addresses the problems of modelling the correlation structure of a set of intervention functions defined on the \DAG of a causal model. We propose the \our model, which is based on a theoretical analysis of the \DAG structure, and allows to share experimental information across interventions while integrating observational and interventional data via \textit{do}-calculus. Our results demonstrate how $\our$ outperforms competing approaches in term of fitting performances. In addition, our experiments show how integrating decision making algorithms with the \our model is crucial when designing optimal experiments as \our accounts for the uncertainty reduction obtained by transferring interventional data. Future work will extend the \our model to allow for transfer of experimental information \emph{across} environments whose \DAG{s} are partially different. In addition, we will focus on combining the proposed framework with a causal discovery algorithm so as to account for uncertainty in the graph structure.

%% file: impact.tex
\section{Broader Impact}
Computing causal effects is an integral part of scientific inquiry, spanning a wide range of questions such as understanding behaviour in online systems, assessing the effect of social policies, or investigation the risk factors for diseases. By combining the theory of causality with machine learning techniques, Causal Machine Learning algorithms have the potential to highly impact society and businesses by answering what-if questions, enabling policy-evaluation and allowing for data-driven decision making in real-world contexts. The algorithm proposed in this paper falls into this category and focuses on addressing causal questions in a fast and accurate way. As shows in the experiments, when used within decision making algorithms, the \our model has the potential to speed up the learning process and to enable optimal experimentation decisions by accounting for the multiple causal connections existing in the process under investigation and their cross-correlation. Our algorithm can be used by practitioners in several domains. For instance, it can be used to learn about the impact of environmental variables on coral calcification \cite{courtney2017environmental} or to analyse the effects of drugs on cancer antigens \cite{ferro2015use}. In terms of methodology, while the \our model represents a step towards a better model for automated decision making, it is based on the crucial assumption of knowing the causal graph. Learning the intervention functions of an incorrect causal graph might lead to incorrect inference and sub-optimal decisions. Therefore, more work needs to be done to account for the uncertainty in the graph structure. 